%% file: main.tex
\documentclass[10pt,twocolumn,letterpaper]{article}

\usepackage[pagenumbers]{cvpr} 

\definecolor{cvprblue}{rgb}{0.21,0.49,0.74}
\usepackage[pagebackref,breaklinks,colorlinks,allcolors=cvprblue]{hyperref}

\def\confName{CVPR}
\def\confYear{2027}

\title{ARISE:  Authentic Reference-Informed Stability Enhancement for Continual  Multimodal Forgery Detection}

\author{
Yuchen Zhang$^{1,3}$ \quad
Yaxiong Wang$^{2,*}$ \quad
Yujiao Wu$^{4}$ \quad
Lianwei Wu$^{5}$ \quad
Qi Zhang$^{3}$ \\
Li Zhu$^{1}$ \quad
Chaofeng Dong$^{6}$ \quad
Le Wu$^{2}$ \quad
Zhedong Zheng$^{7}$ \\[6pt]
{\small $^{1}$School of Software Engineering, Xi'an Jiaotong University} \\
{\small $^{2}$School of Computer and Information Engineering, Hefei University of Technology} \\
{\small
$^{3}$AGI Lab \quad
$^{4}$CSIRO \quad
$^{5}$Northwestern Polytechnical University
} \quad
{\small
$^{6}$Tongji University \quad
$^{7}$University of Macau
} \\
{\small $^{*}$Corresponding author}
}

\usepackage[table]{xcolor}
\usepackage{booktabs}
\usepackage{tabularx}
\usepackage{multirow}
\usepackage{enumitem}
\usepackage{xcolor}
\usepackage{titletoc}

\newlength{\tripanelheight}
\usepackage{adjustbox}

\newlength{\analysispanelheight}
\usepackage{booktabs}
\usepackage{multirow}

\usepackage{pifont}

\usepackage{booktabs}
\usepackage{multirow}
\usepackage[table]{xcolor}
\IfFileExists{fontawesome5.sty}{\usepackage{fontawesome5}}{%
    \newcommand{\faPlay}{\ensuremath{\blacktriangleright}}%
}

\definecolor{GroupGray}{RGB}{242,242,242}
\definecolor{OursBlue}{RGB}{229,241,250}
\definecolor{DividerGray}{RGB}{150,150,150}
\newcommand{\best}[1]{\textbf{#1}}
\newcommand{\second}[1]{\underline{#1}}

\DeclareMathSizes{7.2}{7.2}{5}{5} 

\newcolumntype{Y}{>{\raggedright\arraybackslash}X}

\newcommand{\method}{ARISE}
\newcommand{\bench}{Fable-Bench}

\definecolor{tablegray}{RGB}{252, 251, 242}  
\definecolor{tableblue}{RGB}{235, 243, 255} 
\definecolor{tablesky}{RGB}{230, 230, 250}    

\newcommand{\fatriangle}[2]{\scalebox{0.7}{\rotatebox{#1}{#2}}}

\newcommand{\triRight}{\fatriangle{0}{\faPlay}}

\definecolor{myblue}{RGB}{31, 119, 180}
\definecolor{myred}{RGB}{214, 39, 40}

\begin{document}
\pagestyle{plain}
\twocolumn[
{
\renewcommand\twocolumn[1][]{#1}
\maketitle

\begin{center}
\setlength{\abovecaptionskip}{8pt}
\setlength{\belowcaptionskip}{0pt}
\centering
\includegraphics[
    width=\textwidth,
    height=1.6in,
    keepaspectratio
]{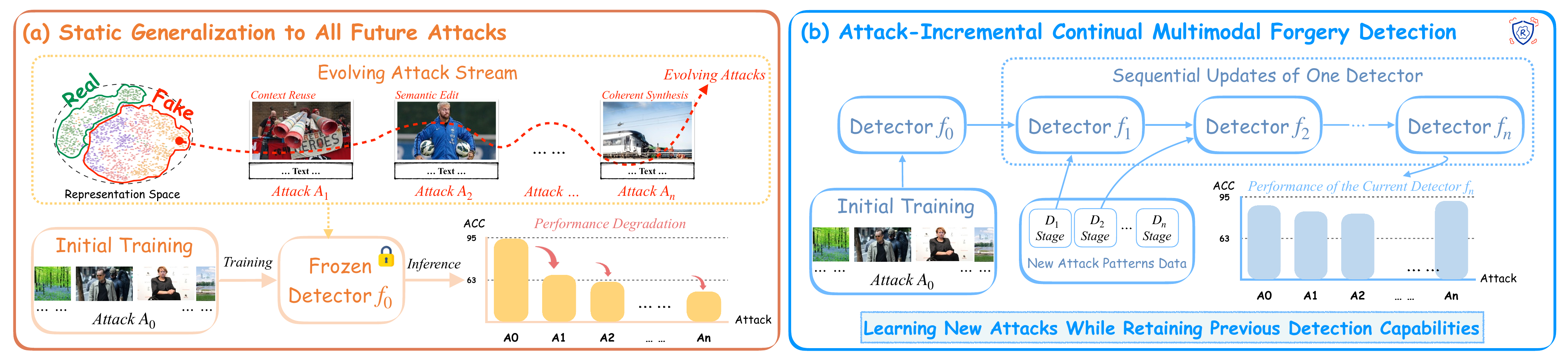}
\vspace{-0.65cm}
\captionof{figure}{Conceptual motivation for attack-incremental forgery detection.
(a) A detector trained once on an initial attack-space snapshot is frozen
after deployment. As new forgery mechanisms emerge, the resulting
distribution shifts alter forensic evidence and could substantially degrade
detection performance.
(b) We instead sequentially update the detector with newly available
labeled attack data. After each stage, the current detector is evaluated on
all attacks observed so far.}
\label{fig:motivation}
\end{center}
}
]

\input{00_abs}

\input{01_introduction}
\input{02_related_work}

\input{05_method}

\input{06_experiments}

\input{08_conclusion}

{
    \small
    \bibliographystyle{ieeenat_fullname}
    \bibliography{main}
}

\end{document}

%% file: 00_abs.tex
\begin{abstract}
Multimodal forgery detection is commonly studied under a static assumption:
training attacks are expected to approximate those encountered after
deployment.
This assumption becomes increasingly fragile as generators, editing systems,
and multimodal large language models continually introduce new forgery
mechanisms and forensic evidence.
We formulate this problem as \emph{attack-incremental continual multimodal
forgery detection}, where the binary real/fake interface remains fixed while
emerging attacks must be learned without forgetting previous ones.
We introduce \bench{}, which organizes attack evolution through complementary
joint-modality, text-side, and image-side protocols.
We further propose \method{}
(\emph{Authentic Reference-Informed Stability Enhancement}), motivated by the
open-ended forged class and the comparatively reusable geometry of authentic
content.
\method{} fits a real-only Gaussian mixture model to projected authentic
replay representations as a compact historical reference.
Meanwhile, the proposed Gaussian-Responsibility-Aware
Stability Preservation (GRASP) derives a sample-adaptive anisotropic metric that strongly constrains
historically stable representation directions while retaining plasticity
elsewhere.
Our Vulnerability-Aware Replay (VAR) further estimates stage-wise vulnerability from authentic-density degradation,
forged-to-authentic boundary confusion, and authentic-mode occupancy drift,
then assigns bounded replay emphasis to fragile stages.
All adaptive signals are obtained from training-time replay statistics, while
inference uses only the original multimodal backbone and binary classification
head.
Experiments across three attack-evolution protocols and multiple detector
architectures demonstrate improved historical retention and final average
performance over representative continual-learning baselines.
\end{abstract}

%% file: 01_introduction.tex
\section{Introduction}
\label{sec:introduction}

Advances in generative artificial intelligence have made it easier to
fabricate image--text news using manipulated images, misleading captions,
and machine-written narratives~\cite{Olan2024_fakeNewsImpact,Zhou2026REVEAL}.
Detecting such content requires examining both individual modalities
and the relationship between them.

Multimodal misinformation detection has progressed from event-level
classification to fine-grained manipulation analysis.
Early methods learn event-invariant representations or detect
out-of-context image reuse~\cite{wang2018eann,aneja2021cosmos}.
Later approaches jointly detect and ground
manipulations~\cite{CVPR_DGM4}, improve cross-modal
alignment~\cite{zhang2024asap}, address coordinated
forgeries~\cite{shen2025samm,zhang2025mdsm}, and incorporate
forensic reasoning~\cite{zhang2026reform}.
These studies improve detection and generalization, but typically
evaluate detectors trained on a fixed collection of attacks.

As generation and editing tools evolve, a deployed detector encounters
forgery mechanisms that were absent during training.
Generalization to these attacks remains important, but cannot be
guaranteed without assumptions relating the training and deployment
distributions~\cite{ye2021theoretical}.
Figure~\ref{fig:motivation} illustrates an alternative:
when labeled examples of a new attack become available, the detector
is updated while preserving its ability to recognize earlier attacks.
Continual learning~\cite{delange2022survey} supports this process
without requiring repeated training on the complete historical dataset.
We study this setting as \emph{attack-incremental continual multimodal
forgery detection}.
Here, an attack refers to a forgery-generation or manipulation mechanism.
The real/fake label space remains fixed, while attack mechanisms and
the evidence needed to detect them change.
This setting is a forensics-specific form of domain-incremental learning,
with shifts driven by emerging attacks.

To study this setting, we introduce \textbf{\bench{}},
short for \emph{Forgery Attack Benchmark for Lifelong Evaluation}.
It provides three protocols for joint-modality, text-side, and
image-side attack evolution.
Each protocol requires a single detector to learn successive attacks
with limited replay memory and retain its performance on previously
seen tasks.

We propose \textbf{\method{}}
(\emph{Authentic Reference-Informed Stability Enhancement}),
which uses historical authentic samples as a reference for continual
updates.
While new attacks introduce diverse forged distributions, authentic
content provides a common reference across these updates.
At each stage, \method{} freezes the preceding detector and fits
a real-only Gaussian mixture model to its projected authentic
replay representations.
GRASP (\emph{Gaussian-Responsibility-Aware Stability Preservation})
uses the mixture responsibilities and variances to constrain
representation changes more strongly along locally low-variance
directions, while allowing greater changes along more variable ones.
Vulnerability-Aware Replay (VAR) complements this constraint by
assigning greater replay weight to historical stages that show
authentic-density degradation, forged-to-authentic boundary confusion,
or changes in authentic-mode occupancy.
Both components use training-time replay statistics.
Inference requires only the multimodal backbone and binary
classification head.

Our contributions are threefold:
\begin{itemize}[leftmargin=*,nosep]
\item We study attack-incremental continual multimodal forgery
detection, where a single real/fake detector learns emerging attacks
while retaining its performance on earlier ones.

\item We introduce \bench{}, which evaluates joint-modality,
text-side, and image-side attack evolution through three
continual-learning protocols.

\item We propose \method{}, combining authentic-reference-based
representation preservation with vulnerability-aware replay.
Experiments on \bench{}, two additional benchmarks, and multiple
backbones show improved adaptation and retention without adding
inference-time components.
\end{itemize}

%% file: 02_related_work.tex
\section{Related Work}
\label{sec:related}

\begin{figure*}[t]
\centering
\includegraphics[width=0.9\linewidth]
{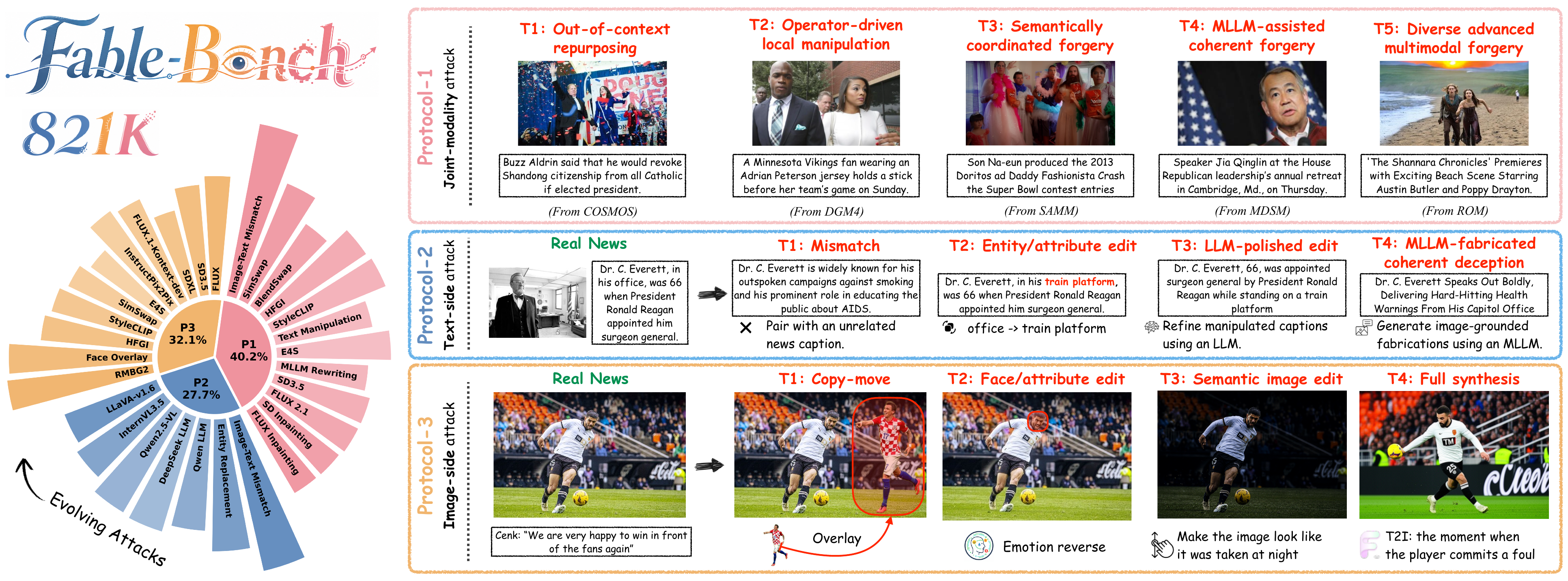}
\vspace{-0.35cm}
\caption{
Overview of \bench{} and its attack-evolution protocols.
Left: 821K image--text pairs spanning 30+ attack mechanisms.
Right: examples of joint-modality (P1), text-side (P2), and
image-side (P3) evolution.
Shared source examples in P2/P3 are used only to illustrate
attack types.
}
\vspace{-0.35cm}
\label{fig:fable_overview}
\end{figure*}

\noindent\textbf{Multimodal Misinformation Detection.}
EANN~\cite{wang2018eann} learns event-invariant multimodal
representations, dEFEND~\cite{shu2019defend} uses news--comment
co-attention, and COSMOS~\cite{aneja2021cosmos} detects
out-of-context image--text pairs.
Later work targets finer-grained manipulations:
DGM4~\cite{CVPR_DGM4} jointly detects and grounds them,
while ASAP~\cite{zhang2024asap} and SAMM~\cite{shen2025samm}
model semantic alignment and coordinated image--text forgeries.
These methods generally assume joint access to attack
distributions during training, whereas \bench{} organizes
sequentially emerging attacks, and \method{} continually
updates a single detector without inference-time branches.

\noindent\textbf{Continual Learning.}
Continual learning mitigates catastrophic
forgetting~\cite{delange2022survey} via architecture expansion
(D-MoLE~\cite{ge2025dynamic}, MoE-CLIP~\cite{yu2024moeclip}),
replay (iCaRL~\cite{rebuffi2017icarl}, LPR~\cite{yoo2024lpr}),
or regularization (MAS~\cite{aljundi2018mas},
C-Flat~\cite{bian2024cflat}).
DAEDCMD~\cite{wang2025daedcmd} studies continual multimodal
misinformation detection across evolving events.
\bench{} focuses on joint-modality, text-side, and image-side
attack evolution.
\method{} combines bounded replay with
authentic-reference-informed stabilization (GRASP and VAR),
without adding inference-time capacity.

%% file: 05_method.tex
\section{Methodology}
\label{sec:method}
\subsection{Problem Setting}
\label{sec:problem_setting}

We study multimodal forgery detection under an
\emph{attack-incremental continual learning} setting.
An input sample is an image--text pair
$x=(I,Q)\in\mathcal{X}$ with a fixed binary label
$y\in\{0,1\}$, where $0$ and $1$ denote authentic and forged
content, respectively.
Training data arrive sequentially as
\begin{equation}
\label{eq:stage_sequence}
\mathcal{D}_{0},\mathcal{D}_{1},\ldots,\mathcal{D}_{T-1},
\end{equation}
where each stage provides a newly available labeled set associated
with an emerging attack distribution.
The label space remains unchanged, but the forged distribution
$p_t(x\mid y=1)$ and the evidence separating forged content from
authentic content could vary across stages.

At stage $t$, the learner can access the current training set
$\mathcal{D}^{\mathrm{tr}}_t$, a bounded replay memory
$\mathcal{M}_{t-1}$, and the detector obtained after stage $t-1$.
The complete historical training data are unavailable.
After each update, a single detector is evaluated on all observed stages without stage identity, requiring it to learn emerging forensic evidence while retaining previously acquired knowledge.

\subsection{\bench{}}
\label{sec:fable}

To evaluate detection under evolving attacks, we construct \bench{} (\emph{Forgery Attack Benchmark for Lifelong Evaluation}), which organizes successive attack stages rather than another static forgery collection. All stages share the same real/fake semantics, and the detector is evaluated on all seen attacks after each update. As in Fig.~\ref{fig:fable_overview}, \bench{} comprises three protocols covering joint-modality, text-side, and image-side evolution.

\noindent\textbf{Protocol-1: Joint-Modality Attack Evolution.}
This protocol combines changes in attack mechanisms and data sources
through the sequence
\textsc{COSMOS}~\cite{aneja2021cosmos}
$\rightarrow$
\textsc{DGM4}~\cite{CVPR_DGM4}
$\rightarrow$
\textsc{SAMM}~\cite{shen2025samm}
$\rightarrow$
\textsc{MDSM}~\cite{zhang2025mdsm}
$\rightarrow$
\textsc{ROM}~\cite{zhang2026reform}.
The five stages cover out-of-context repurposing, local manipulation,
semantically coordinated forgeries, MLLM-assisted fabrication,
and diverse attacks.

\noindent\textbf{Protocol-2: Text-Side Attack Evolution.}
With images kept authentic, textual attacks progress from random mismatch
(T1) and targeted entity or attribute modification (T2) to
language-model-polished manipulation (T3) and image-conditioned coherent
deception (T4). This protocol tests whether coarse inconsistency cues can be
retained while increasingly subtle semantic evidence is learned.

\noindent\textbf{Protocol-3: Image-Side Attack Evolution.}
With accompanying text semantics controlled, image attacks progress from
copy--move manipulation (T1) and face or attribute editing (T2) to
instruction-guided semantic editing (T3) and caption-consistent full-image
synthesis (T4), shifting the dominant evidence from local forensic traces to
global synthetic characteristics.

Overall, \bench{} contains \textbf{821,387} image--text pairs
spanning three protocols and over 30 manipulation models and
attack mechanisms.
We deduplicate P1 across stages and construct P2/P3 from shared
news corpora with disjoint source items.

\subsection{\method{}: Authentic Reference-Informed Stability Enhancement}
\label{sec:arise}

\begin{figure*}[t]
    \centering
    \includegraphics[width=0.95\linewidth]
    {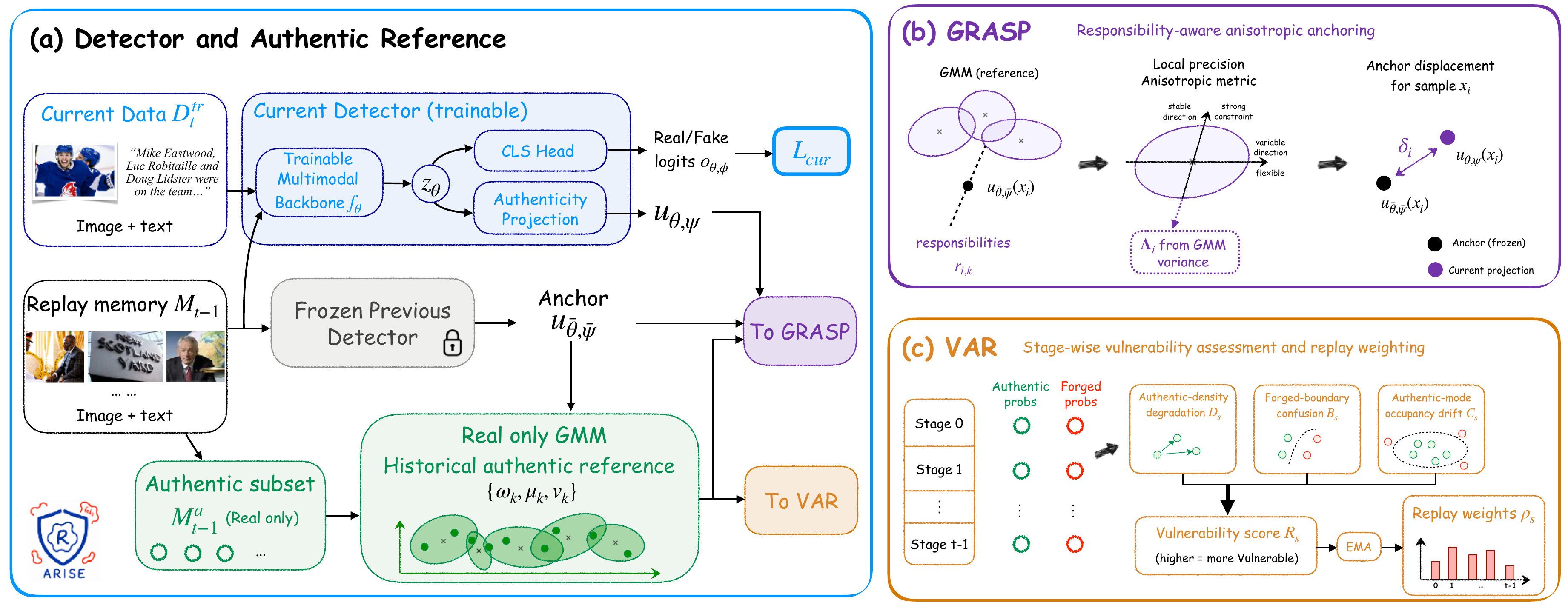}
    \vspace{-0.1cm}
    \caption{
    Overview of \method{}. \textbf{(a)} A frozen previous detector and a real-only Gaussian mixture provide historical anchors during continual updates. \textbf{(b)} GRASP derives a responsibility-aware anisotropic preservation metric from the local geometry of the authentic reference. \textbf{(c)} VAR estimates stage-wise vulnerability and reweights historical replay accordingly. All auxiliary components are removed at inference.
    }

    \label{fig:arise_overview}
    \vspace{-0.35cm}
\end{figure*}

At stage $t$, the detector learns the newly available attack distribution in
$\mathcal{D}^{\mathrm{tr}}_t$ together with a bounded replay memory
$\mathcal{M}_{t-1}$ maintained by standard Experience Replay
(ER)~\cite{ER_replay}.
However, replay classification neither explicitly preserves historical
representation geometry nor accounts for the heterogeneous and time-varying
vulnerability of historical stages, while uniform feature constraints may
limit adaptation to new attacks.
We therefore use replayed authentic content as a compact historical
reference.
Based on this shared reference, GRASP performs sample-adaptive anisotropic
stabilization, while VAR assigns bounded replay emphasis to vulnerable
historical stages.

As shown in Fig.~\ref{fig:arise_overview}a, let $f_\theta$ denote the multimodal
backbone, and let $h_{\mathrm{cls},\phi}$ and
$h_{\mathrm{auth},\psi}$ denote the binary classification head and the
auxiliary authenticity projection, respectively.
For an input $x$, the backbone produces
$\mathbf{z}_\theta(x)=f_\theta(x)$, on which the two heads operate:
\begin{equation}
\label{eq:arise_heads}
\begin{aligned}
\mathbf{o}_{\theta,\phi}(x)
&= h_{\mathrm{cls},\phi}\!\left(\mathbf{z}_\theta(x)\right),\\
\mathbf{u}_{\theta,\psi}(x)
&= h_{\mathrm{auth},\psi}\!\left(\mathbf{z}_\theta(x)\right).
\end{aligned}
\end{equation}
Here, $\mathbf{o}_{\theta,\phi}(x)\in\mathbb{R}^{2}$ contains the
authentic and forged logits, while
$\mathbf{u}_{\theta,\psi}(x)\in\mathbb{R}^{d}$ is a projected
representation used only for authentic-reference modeling and the proposed
continual-learning objectives.
During initial training, the projection shares parameters with the
classification branch; these parameters are then copied into an
independent projection head optimized through GRASP.

At the beginning of stage $t>0$, the current detector is initialized from the
detector obtained at stage $t-1$.
We freeze copies of its backbone and authenticity projection, denoted by
$f_{\bar{\theta}}$ and $h_{\mathrm{auth},\bar{\psi}}$, respectively.
Because the current and frozen branches are identical at initialization,
\begin{equation}
\label{eq:frozen_anchor_initialization}
\mathbf{u}_{\theta,\psi}(x)
=
\mathbf{u}_{\bar{\theta},\bar{\psi}}(x).
\end{equation}
Only the current branch is subsequently updated, while the frozen branch
provides historical anchor representations before learning the newly
arriving attack.

Let $\mathcal{M}^{\mathrm{a}}_{t-1}$ denote the retained historical authentic
samples available at the beginning of stage $t$.
Then we use the frozen detector to extract their projected anchor representations
and fit a Gaussian Mixture Model (GMM):
\begin{equation}
\label{eq:authentic_gmm}
q_{\mathrm{ref}}(\mathbf{u})
=
\sum_{k=1}^{K_t}
\omega_k
\mathcal{N}\!\left(
\mathbf{u};
\boldsymbol{\mu}_k,
\operatorname{diag}(\mathbf{v}_k)
\right).
\end{equation}
Here, $q_{\mathrm{ref}}$ denotes the historical authentic-reference
density, and $\{\omega_k,\boldsymbol{\mu}_k,\mathbf{v}_k\}_{k=1}^{K_t}$
are the mixture weights, means, and diagonal variances.
The reference is fitted once per stage using stage-balanced
weighted expectation--maximization and remains fixed throughout
the stage.

To accommodate the growing diversity of historical authentic content, the
component budget increases with the number of historical stages:
\begin{equation}
\label{eq:gmm_component_budget}
K_t
=
\min\!\left(
4t,\,
20,\,
\left|\mathcal{M}^{\mathrm{a}}_{t-1}\right|
\right).
\end{equation}
This design progressively expands the reference capacity while avoiding
excessive fragmentation under bounded replay memory.

The resulting GMM serves as a shared historical reference for the two
subsequent components of \method{}.
GRASP uses its local precision geometry to identify representation directions
that should remain stable. VAR uses its components and stage-specific
density statistics to identify historical stages requiring stronger replay
protection.

\subsection{GRASP: Gaussian-Responsibility-Aware Stability Preservation}
\label{sec:grasp}

GRASP addresses the lack of geometry-aware preservation in standard replay.
It derives a sample-dependent anisotropic preservation metric from the
authentic GMM, selectively constraining historically stable directions while
retaining flexibility along naturally variable ones.
Consider a historical sample $x_i$ drawn from the replay mini-batch
$\mathcal{B}^{\mathrm{rep}}$.
Let $\mathbf{a}_i = \mathbf{u}_{\bar{\theta},\bar{\psi}}(x_i)$
be its frozen anchor representation.
Its posterior responsibility for Gaussian component $k$ is
\begin{equation}
\label{eq:grasp_responsibility}
r_{i,k}
=
\frac{
\omega_k
\mathcal{N}\!\left(
\mathbf{a}_i;
\boldsymbol{\mu}_k,
\operatorname{diag}(\mathbf{v}_k)
\right)
}{
\sum_{j=1}^{K_t}
\omega_j
\mathcal{N}\!\left(
\mathbf{a}_i;
\boldsymbol{\mu}_j,
\operatorname{diag}(\mathbf{v}_j)
\right)
}.
\end{equation}
Thus, $r_{i,k}$ quantifies how strongly the historical representation of
$x_i$ is associated with component $k$, and
$\sum_{k=1}^{K_t}r_{i,k}=1$.

For each Gaussian component, inverse diagonal variance gives
higher weight to feature directions with less variation among
historical authentic samples.
Here, low variance reflects local consistency among historical
authentic representations, but does not necessarily imply
discriminative importance.
GRASP combines these component-wise precisions using
sample-specific responsibilities:
\begin{equation}
\label{eq:grasp_precision}
\boldsymbol{\lambda}_i
=
\operatorname{clip}\!\left(
\sum_{k=1}^{K_t}
\frac{r_{i,k}}{\max(\mathbf{v}_k,\varepsilon_{\mathrm{var}})},
0,\lambda_{\max}
\right).
\end{equation}
We set $\boldsymbol{\Lambda}_i=\operatorname{diag}(\boldsymbol{\lambda}_i)$.
All vector operations in Eq.~\eqref{eq:grasp_precision} are element-wise.
The variance floor $\varepsilon_{\mathrm{var}}>0$ prevents numerical
instability for small component variances, while $\lambda_{\max}$ limits
excessively large preservation gradients.

The current representation displacement is
\begin{equation}
\boldsymbol{\delta}_i
=
\mathbf{u}_{\theta,\psi}(x_i)
-
\operatorname{sg}\!\left[
\mathbf{u}_{\bar{\theta},\bar{\psi}}(x_i)
\right],
\end{equation}
where $\operatorname{sg}[\cdot]$ denotes stop-gradient.
The GRASP objective is then
\begin{equation}
\label{eq:grasp_loss}
\mathcal{L}_{\mathrm{GRASP}}
=
\mathbb{E}_{x_i\sim\mathcal{B}^{\mathrm{rep}}}
\left[
\boldsymbol{\delta}_i^{\top}
\boldsymbol{\Lambda}_i
\boldsymbol{\delta}_i
\right].
\end{equation}
The frozen representations, posterior responsibilities, local precisions,
and GMM parameters are detached from gradient computation.
Gradients flow through $\mathbf{u}_{\theta,\psi}(x_i)$ to both
the current backbone and the authenticity projection.
Both authentic and forged replay samples participate in
Eq.~\eqref{eq:grasp_loss}.
Although the metric is estimated from authentic geometry, it constrains
representation-update directions rather than sample authenticity.
Hence it applies to any sample whose gradient updates the current detector. 


\subsection{VAR: Vulnerability-Aware Replay}
\label{sec:var}

GRASP stabilizes historical representations, but different stages 
exhibit unequal vulnerability to forgetting.
Such vulnerability depends not only on stage age, but also on the
interaction between current optimization and historical attack
distributions.
As shown in Fig.~\ref{fig:arise_overview}c, VAR periodically estimates
stage-level vulnerability from accumulated diagnostic signals and assigns
bounded replay weights to vulnerable stages.

\noindent\textbf{Stage Vulnerability Assessment.} For each historical stage $s<t$, VAR constructs fixed authentic and forged
probe subsets at the beginning of stage $t$, where
$\operatorname{stage}(x)$ denotes the source-stage index of replay sample $x$:
\begin{equation}
\mathcal{P}^{\mathrm{a}}_s
\subseteq
\left\{
x\in\mathcal{M}_{t-1}
\,\middle|\,
\operatorname{stage}(x)=s,\ y=0
\right\},
\end{equation}
\begin{equation}
\mathcal{P}^{\mathrm{f}}_s
\subseteq
\left\{
x\in\mathcal{M}_{t-1}
\,\middle|\,
\operatorname{stage}(x)=s,\ y=1
\right\}.
\end{equation}
These subsets are used for periodic sample-level diagnostics, whereas all retained authentic replay samples from stage $s$ are used to construct the more stable stage occupancy profiles.

VAR then constructs a target component profile for each historical stage:
\begin{equation}
\label{eq:var_target_profile}
\boldsymbol{\pi}_s
=
\frac{
\mathbf{1}
+
\sum_{x\in\mathcal{M}^{\mathrm{a}}_{t-1,s}}
\mathbf{r}\!\left(
\mathbf{u}_{\bar{\theta},\bar{\psi}}(x)
\right)
}{
K_t
+
|\mathcal{M}^{\mathrm{a}}_{t-1,s}|
},
\end{equation}
where $\mathcal{M}^{\mathrm{a}}_{t-1,s}$ contains the authentic replay samples
from stage $s$, and
$\mathbf{r}(\mathbf{u})
=
[r_1(\mathbf{u}),\ldots,r_{K_t}(\mathbf{u})]^\top$
is the responsibility vector under the frozen global GMM in
Eq.~\eqref{eq:authentic_gmm}.
Thus, $\pi_{s,k}$ denotes the target occupancy of component $k$ for stage $s$.

\begin{table*}[t]
\centering
\caption{
Accuracy (\%) under the joint-modality (P1), text-side (P2), and image-side
(P3) attack-evolution protocols. Results are reported as mean $\pm$ standard
deviation over four runs.
\colorbox[HTML]{F7C9CC}{Red} and \colorbox[HTML]{D3E1F6}{Blue} indicate the
best and second-best results, respectively, with MLLM and reference baselines
excluded from ranking.
}
\label{tab:main_results}
\vspace{-0.3cm}
\setlength{\tabcolsep}{3.0pt}
\renewcommand{\arraystretch}{1.20}

\resizebox{0.99\textwidth}{!}{%
\begin{tabular}{
    l
    *{6}{c}
    !{\color{DividerGray}\vrule width 0.7pt}
    *{5}{c}
    !{\color{DividerGray}\vrule width 0.7pt}
    *{5}{c}
}
\toprule
\multirow{2}{*}{Method}
& \multicolumn{6}{c!{\color{DividerGray}\vrule width 0.7pt}}{Joint-Modality Protocol (P1)}
& \multicolumn{5}{c!{\color{DividerGray}\vrule width 0.7pt}}{Text-Side Protocol (P2)}
& \multicolumn{5}{c}{Image-Side Protocol (P3)}
\\
\cmidrule(lr){2-7}
\cmidrule(lr){8-12}
\cmidrule(lr){13-17}
& T1 & T2 & T3 & T4 & T5 & Avg.
& T1 & T2 & T3 & T4 & Avg.
& T1 & T2 & T3 & T4 & Avg.
\\
\midrule

\rowcolor{GroupGray}
\multicolumn{17}{l}{\textit{MLLM-based Baselines}} \\

GPT-5.4-mini~\cite{openai2026gpt54mini}
& 92.24 & 64.05 & 69.51 & 62.26 & 62.08 & 70.03
& 91.70 & 64.08 & 60.36 & 59.32 & 68.86
& 61.15 & 58.37 & 68.04 & 73.33 & 65.22
\\

Gemini-3-Flash~\cite{googledeepmind2025gemini3flash}
& 93.16 & 65.23 & 70.79 & 63.50 & 63.09 & 71.15
& 92.19 & 64.97 & 61.40 & 60.31 & 69.72
& 62.86 & 60.10 & 70.64 & 75.42 & 67.25
\\

Qwen3.5-Plus~\cite{qwen2026qwen35}
& 90.53 & 61.18 & 66.72 & 59.48 & 59.39 & 67.46
& 89.98 & 61.20 & 56.73 & 55.92 & 65.96
& 57.64 & 54.91 & 63.88 & 69.05 & 61.37
\\

InternVL3.5-30B~\cite{wang2025internvl35}
& 88.92 & 59.36 & 63.97 & 57.10 & 57.50 & 65.37
& 87.54 & 58.29 & 54.04 & 52.90 & 63.19
& 59.69 & 57.01 & 65.65 & 71.15 & 63.37
\\

LLaVA-v1.6-7B~\cite{liu2024llavanext}
& 87.78 & 55.12 & 59.19 & 53.26 & 53.83 & 61.83
& 86.53 & 54.32 & 51.21 & 50.02 & 60.52
& 53.69 & 51.87 & 59.47 & 64.82 & 57.46
\\

\midrule

\rowcolor{GroupGray}
\multicolumn{17}{l}{\textit{Reference Baselines}} \\

Joint Training
& 93.07 & 80.34 & 90.54 & 95.69 & 94.70 & 90.87
& 84.14 & 82.96 & 86.45 & 93.58 & 86.78
& 97.90 & 81.84 & 91.71 & 94.37 & 91.45
\\

Naive Fine-tuning
& 50.04 & 61.11 & 54.10 & 80.84 & 92.07 & 67.63
& 50.21 & 51.26 & 73.84 & 99.11 & 68.60
& 68.07 & 45.77 & 67.42 & 96.95 & 69.55
\\

Experience Replay
& 70.83 & 74.03 & 80.46 & 90.25 & 92.77 & 81.67
& 68.35 & 71.92 & 75.82 & 94.60 & 77.67
& 80.33 & 78.38 & 80.44 & 95.08 & 83.56
\\

\midrule

\rowcolor{GroupGray}
\multicolumn{17}{l}{\textit{Architecture-based Methods}} \\

MoE-CLIP~\cite{yu2024moeclip}
& \shortstack[c]{76.89\\[-1.2pt]{\tiny\color{black!38}$\pm0.10$}} & \shortstack[c]{68.18\\[-1.2pt]{\tiny\color{black!38}$\pm0.10$}} & \shortstack[c]{76.40\\[-1.2pt]{\tiny\color{black!38}$\pm0.16$}} & \shortstack[c]{86.41\\[-1.2pt]{\tiny\color{black!38}$\pm0.20$}} & \shortstack[c]{88.21\\[-1.2pt]{\tiny\color{black!38}$\pm0.06$}} & \shortstack[c]{79.22\\[-1.2pt]{\tiny\color{black!38}$\pm0.08$}}
& \shortstack[c]{65.72\\[-1.2pt]{\tiny\color{black!38}$\pm0.10$}} & \shortstack[c]{68.73\\[-1.2pt]{\tiny\color{black!38}$\pm0.13$}} & \shortstack[c]{75.90\\[-1.2pt]{\tiny\color{black!38}$\pm0.09$}} & \shortstack[c]{84.83\\[-1.2pt]{\tiny\color{black!38}$\pm0.15$}} & \shortstack[c]{73.79\\[-1.2pt]{\tiny\color{black!38}$\pm0.07$}}
& \shortstack[c]{87.55\\[-1.2pt]{\tiny\color{black!38}$\pm0.14$}} & \shortstack[c]{80.88\\[-1.2pt]{\tiny\color{black!38}$\pm0.08$}} & \shortstack[c]{76.91\\[-1.2pt]{\tiny\color{black!38}$\pm0.18$}} & \shortstack[c]{93.64\\[-1.2pt]{\tiny\color{black!38}$\pm0.11$}} & \shortstack[c]{84.74\\[-1.2pt]{\tiny\color{black!38}$\pm0.09$}}
\\

D-MoLE~\cite{ge2025dynamic}
& \shortstack[c]{80.35\\[-1.2pt]{\tiny\color{black!38}$\pm0.08$}} & \cellcolor[HTML]{D3E1F6}\shortstack[c]{73.22\\[-1.2pt]{\tiny\color{black!38}$\pm0.08$}} & \shortstack[c]{75.40\\[-1.2pt]{\tiny\color{black!38}$\pm0.13$}} & \shortstack[c]{89.13\\[-1.2pt]{\tiny\color{black!38}$\pm0.10$}} & \shortstack[c]{92.06\\[-1.2pt]{\tiny\color{black!38}$\pm0.12$}} & \shortstack[c]{82.03\\[-1.2pt]{\tiny\color{black!38}$\pm0.03$}}
& \shortstack[c]{68.84\\[-1.2pt]{\tiny\color{black!38}$\pm0.12$}} & \shortstack[c]{72.97\\[-1.2pt]{\tiny\color{black!38}$\pm0.17$}} & \shortstack[c]{75.85\\[-1.2pt]{\tiny\color{black!38}$\pm0.10$}} & \shortstack[c]{82.35\\[-1.2pt]{\tiny\color{black!38}$\pm0.17$}} & \shortstack[c]{75.00\\[-1.2pt]{\tiny\color{black!38}$\pm0.08$}}
& \shortstack[c]{90.81\\[-1.2pt]{\tiny\color{black!38}$\pm0.21$}} & \shortstack[c]{82.57\\[-1.2pt]{\tiny\color{black!38}$\pm0.10$}} & \cellcolor[HTML]{D3E1F6}\shortstack[c]{83.84\\[-1.2pt]{\tiny\color{black!38}$\pm0.19$}} & \shortstack[c]{94.55\\[-1.2pt]{\tiny\color{black!38}$\pm0.15$}} & \shortstack[c]{87.94\\[-1.2pt]{\tiny\color{black!38}$\pm0.09$}}
\\

DAEDCMD~\cite{wang2025daedcmd}
& \shortstack[c]{75.50\\[-1.2pt]{\tiny\color{black!38}$\pm0.18$}} & \shortstack[c]{66.77\\[-1.2pt]{\tiny\color{black!38}$\pm0.17$}} & \shortstack[c]{77.96\\[-1.2pt]{\tiny\color{black!38}$\pm0.13$}} & \shortstack[c]{89.12\\[-1.2pt]{\tiny\color{black!38}$\pm0.15$}} & \shortstack[c]{92.25\\[-1.2pt]{\tiny\color{black!38}$\pm0.17$}} & \shortstack[c]{80.32\\[-1.2pt]{\tiny\color{black!38}$\pm0.08$}}
& \shortstack[c]{69.97\\[-1.2pt]{\tiny\color{black!38}$\pm0.10$}} & \shortstack[c]{71.87\\[-1.2pt]{\tiny\color{black!38}$\pm0.18$}} & \shortstack[c]{76.83\\[-1.2pt]{\tiny\color{black!38}$\pm0.18$}} & \shortstack[c]{81.71\\[-1.2pt]{\tiny\color{black!38}$\pm0.13$}} & \shortstack[c]{75.10\\[-1.2pt]{\tiny\color{black!38}$\pm0.09$}}
& \shortstack[c]{94.59\\[-1.2pt]{\tiny\color{black!38}$\pm0.19$}} & \shortstack[c]{82.02\\[-1.2pt]{\tiny\color{black!38}$\pm0.10$}} & \shortstack[c]{78.88\\[-1.2pt]{\tiny\color{black!38}$\pm0.11$}} & \shortstack[c]{93.87\\[-1.2pt]{\tiny\color{black!38}$\pm0.13$}} & \shortstack[c]{87.34\\[-1.2pt]{\tiny\color{black!38}$\pm0.07$}}
\\

\midrule

\rowcolor{GroupGray}
\multicolumn{17}{l}{\textit{Replay-based Methods}} \\

iCaRL~\cite{rebuffi2017icarl}
& \shortstack[c]{66.81\\[-1.2pt]{\tiny\color{black!38}$\pm0.15$}} & \shortstack[c]{70.55\\[-1.2pt]{\tiny\color{black!38}$\pm0.04$}} & \cellcolor[HTML]{D3E1F6}\shortstack[c]{81.26\\[-1.2pt]{\tiny\color{black!38}$\pm0.22$}} & \shortstack[c]{88.84\\[-1.2pt]{\tiny\color{black!38}$\pm0.08$}} & \shortstack[c]{90.11\\[-1.2pt]{\tiny\color{black!38}$\pm0.11$}} & \shortstack[c]{79.51\\[-1.2pt]{\tiny\color{black!38}$\pm0.07$}}
& \shortstack[c]{68.61\\[-1.2pt]{\tiny\color{black!38}$\pm0.13$}} & \shortstack[c]{62.75\\[-1.2pt]{\tiny\color{black!38}$\pm0.13$}} & \shortstack[c]{81.42\\[-1.2pt]{\tiny\color{black!38}$\pm0.13$}} & \shortstack[c]{93.81\\[-1.2pt]{\tiny\color{black!38}$\pm0.13$}} & \shortstack[c]{76.65\\[-1.2pt]{\tiny\color{black!38}$\pm0.07$}}
& \shortstack[c]{89.57\\[-1.2pt]{\tiny\color{black!38}$\pm0.18$}} & \shortstack[c]{78.49\\[-1.2pt]{\tiny\color{black!38}$\pm0.07$}} & \shortstack[c]{77.77\\[-1.2pt]{\tiny\color{black!38}$\pm0.09$}} & \shortstack[c]{95.29\\[-1.2pt]{\tiny\color{black!38}$\pm0.16$}} & \shortstack[c]{85.28\\[-1.2pt]{\tiny\color{black!38}$\pm0.10$}}
\\

LPR~\cite{yoo2024lpr}
& \shortstack[c]{81.05\\[-1.2pt]{\tiny\color{black!38}$\pm0.16$}} & \shortstack[c]{69.72\\[-1.2pt]{\tiny\color{black!38}$\pm0.14$}} & \shortstack[c]{81.10\\[-1.2pt]{\tiny\color{black!38}$\pm0.14$}} & \cellcolor[HTML]{D3E1F6}\shortstack[c]{92.56\\[-1.2pt]{\tiny\color{black!38}$\pm0.12$}} & \shortstack[c]{94.70\\[-1.2pt]{\tiny\color{black!38}$\pm0.09$}} & \cellcolor[HTML]{D3E1F6}\shortstack[c]{83.83\\[-1.2pt]{\tiny\color{black!38}$\pm0.04$}}
& \cellcolor[HTML]{D3E1F6}\shortstack[c]{71.18\\[-1.2pt]{\tiny\color{black!38}$\pm0.16$}} & \shortstack[c]{73.80\\[-1.2pt]{\tiny\color{black!38}$\pm0.09$}} & \shortstack[c]{80.23\\[-1.2pt]{\tiny\color{black!38}$\pm0.02$}} & \shortstack[c]{92.16\\[-1.2pt]{\tiny\color{black!38}$\pm0.16$}} & \cellcolor[HTML]{D3E1F6}\shortstack[c]{79.34\\[-1.2pt]{\tiny\color{black!38}$\pm0.07$}}
& \cellcolor[HTML]{D3E1F6}\shortstack[c]{97.07\\[-1.2pt]{\tiny\color{black!38}$\pm0.12$}} & \shortstack[c]{83.07\\[-1.2pt]{\tiny\color{black!38}$\pm0.13$}} & \shortstack[c]{80.57\\[-1.2pt]{\tiny\color{black!38}$\pm0.16$}} & \shortstack[c]{95.24\\[-1.2pt]{\tiny\color{black!38}$\pm0.14$}} & \cellcolor[HTML]{D3E1F6}\shortstack[c]{88.99\\[-1.2pt]{\tiny\color{black!38}$\pm0.07$}}
\\

aL-SAR~\cite{seo2025alsar}
& \cellcolor[HTML]{F7C9CC}\shortstack[c]{81.96\\[-1.2pt]{\tiny\color{black!38}$\pm0.10$}} & \shortstack[c]{68.29\\[-1.2pt]{\tiny\color{black!38}$\pm0.11$}} & \shortstack[c]{79.54\\[-1.2pt]{\tiny\color{black!38}$\pm0.15$}} & \shortstack[c]{92.48\\[-1.2pt]{\tiny\color{black!38}$\pm0.20$}} & \shortstack[c]{94.92\\[-1.2pt]{\tiny\color{black!38}$\pm0.15$}} & \shortstack[c]{83.44\\[-1.2pt]{\tiny\color{black!38}$\pm0.03$}}
& \shortstack[c]{70.82\\[-1.2pt]{\tiny\color{black!38}$\pm0.05$}} & \shortstack[c]{72.78\\[-1.2pt]{\tiny\color{black!38}$\pm0.17$}} & \cellcolor[HTML]{D3E1F6}\shortstack[c]{81.82\\[-1.2pt]{\tiny\color{black!38}$\pm0.12$}} & \shortstack[c]{90.95\\[-1.2pt]{\tiny\color{black!38}$\pm0.14$}} & \shortstack[c]{79.09\\[-1.2pt]{\tiny\color{black!38}$\pm0.03$}}
& \shortstack[c]{96.55\\[-1.2pt]{\tiny\color{black!38}$\pm0.08$}} & \shortstack[c]{82.31\\[-1.2pt]{\tiny\color{black!38}$\pm0.10$}} & \shortstack[c]{80.09\\[-1.2pt]{\tiny\color{black!38}$\pm0.10$}} & \shortstack[c]{94.26\\[-1.2pt]{\tiny\color{black!38}$\pm0.13$}} & \shortstack[c]{88.30\\[-1.2pt]{\tiny\color{black!38}$\pm0.06$}}
\\

\midrule

\rowcolor{GroupGray}
\multicolumn{17}{l}{\textit{Regularization-based Methods}} \\

MAS~\cite{aljundi2018mas}
& \shortstack[c]{70.19\\[-1.2pt]{\tiny\color{black!38}$\pm0.06$}} & \shortstack[c]{68.29\\[-1.2pt]{\tiny\color{black!38}$\pm0.11$}} & \shortstack[c]{80.99\\[-1.2pt]{\tiny\color{black!38}$\pm0.05$}} & \shortstack[c]{89.07\\[-1.2pt]{\tiny\color{black!38}$\pm0.16$}} & \shortstack[c]{90.08\\[-1.2pt]{\tiny\color{black!38}$\pm0.08$}} & \shortstack[c]{79.72\\[-1.2pt]{\tiny\color{black!38}$\pm0.04$}}
& \shortstack[c]{65.85\\[-1.2pt]{\tiny\color{black!38}$\pm0.16$}} & \shortstack[c]{68.79\\[-1.2pt]{\tiny\color{black!38}$\pm0.12$}} & \shortstack[c]{76.72\\[-1.2pt]{\tiny\color{black!38}$\pm0.11$}} & \cellcolor[HTML]{F7C9CC}\shortstack[c]{96.38\\[-1.2pt]{\tiny\color{black!38}$\pm0.19$}} & \shortstack[c]{76.93\\[-1.2pt]{\tiny\color{black!38}$\pm0.11$}}
& \shortstack[c]{88.46\\[-1.2pt]{\tiny\color{black!38}$\pm0.22$}} & \shortstack[c]{77.56\\[-1.2pt]{\tiny\color{black!38}$\pm0.12$}} & \shortstack[c]{79.38\\[-1.2pt]{\tiny\color{black!38}$\pm0.21$}} & \shortstack[c]{93.71\\[-1.2pt]{\tiny\color{black!38}$\pm0.15$}} & \shortstack[c]{84.78\\[-1.2pt]{\tiny\color{black!38}$\pm0.09$}}
\\

C-Flat~\cite{bian2024cflat}
& \shortstack[c]{76.01\\[-1.2pt]{\tiny\color{black!38}$\pm0.11$}} & \shortstack[c]{69.63\\[-1.2pt]{\tiny\color{black!38}$\pm0.04$}} & \shortstack[c]{77.36\\[-1.2pt]{\tiny\color{black!38}$\pm0.07$}} & \shortstack[c]{90.16\\[-1.2pt]{\tiny\color{black!38}$\pm0.17$}} & \shortstack[c]{92.17\\[-1.2pt]{\tiny\color{black!38}$\pm0.09$}} & \shortstack[c]{81.06\\[-1.2pt]{\tiny\color{black!38}$\pm0.05$}}
& \shortstack[c]{69.70\\[-1.2pt]{\tiny\color{black!38}$\pm0.15$}} & \shortstack[c]{68.42\\[-1.2pt]{\tiny\color{black!38}$\pm0.17$}} & \shortstack[c]{79.61\\[-1.2pt]{\tiny\color{black!38}$\pm0.10$}} & \shortstack[c]{92.62\\[-1.2pt]{\tiny\color{black!38}$\pm0.11$}} & \shortstack[c]{77.59\\[-1.2pt]{\tiny\color{black!38}$\pm0.09$}}
& \shortstack[c]{95.68\\[-1.2pt]{\tiny\color{black!38}$\pm0.08$}} & \shortstack[c]{82.60\\[-1.2pt]{\tiny\color{black!38}$\pm0.11$}} & \shortstack[c]{80.86\\[-1.2pt]{\tiny\color{black!38}$\pm0.09$}} & \shortstack[c]{94.70\\[-1.2pt]{\tiny\color{black!38}$\pm0.08$}} & \shortstack[c]{88.46\\[-1.2pt]{\tiny\color{black!38}$\pm0.05$}}
\\

ZAF~\cite{gao2024zaf}
& \shortstack[c]{80.05\\[-1.2pt]{\tiny\color{black!38}$\pm0.17$}} & \shortstack[c]{70.23\\[-1.2pt]{\tiny\color{black!38}$\pm0.19$}} & \shortstack[c]{78.79\\[-1.2pt]{\tiny\color{black!38}$\pm0.15$}} & \shortstack[c]{92.34\\[-1.2pt]{\tiny\color{black!38}$\pm0.15$}} & \cellcolor[HTML]{F7C9CC}\shortstack[c]{95.49\\[-1.2pt]{\tiny\color{black!38}$\pm0.05$}} & \shortstack[c]{83.38\\[-1.2pt]{\tiny\color{black!38}$\pm0.06$}}
& \shortstack[c]{70.97\\[-1.2pt]{\tiny\color{black!38}$\pm0.08$}} & \cellcolor[HTML]{F7C9CC}\shortstack[c]{74.76\\[-1.2pt]{\tiny\color{black!38}$\pm0.08$}} & \shortstack[c]{80.53\\[-1.2pt]{\tiny\color{black!38}$\pm0.15$}} & \shortstack[c]{83.69\\[-1.2pt]{\tiny\color{black!38}$\pm0.16$}} & \shortstack[c]{77.49\\[-1.2pt]{\tiny\color{black!38}$\pm0.09$}}
& \shortstack[c]{95.33\\[-1.2pt]{\tiny\color{black!38}$\pm0.09$}} & \cellcolor[HTML]{D3E1F6}\shortstack[c]{83.34\\[-1.2pt]{\tiny\color{black!38}$\pm0.09$}} & \shortstack[c]{81.12\\[-1.2pt]{\tiny\color{black!38}$\pm0.11$}} & \cellcolor[HTML]{D3E1F6}\shortstack[c]{95.84\\[-1.2pt]{\tiny\color{black!38}$\pm0.14$}} & \shortstack[c]{88.91\\[-1.2pt]{\tiny\color{black!38}$\pm0.07$}}
\\

\midrule

\textbf{\method{} (Ours)}
& \cellcolor[HTML]{D3E1F6}\shortstack[c]{81.21\\[-1.2pt]{\tiny\color{black!38}$\pm0.03$}} & \cellcolor[HTML]{F7C9CC}\shortstack[c]{74.22\\[-1.2pt]{\tiny\color{black!38}$\pm0.12$}} & \cellcolor[HTML]{F7C9CC}\shortstack[c]{82.72\\[-1.2pt]{\tiny\color{black!38}$\pm0.13$}} & \cellcolor[HTML]{F7C9CC}\shortstack[c]{93.73\\[-1.2pt]{\tiny\color{black!38}$\pm0.11$}} & \cellcolor[HTML]{D3E1F6}\shortstack[c]{95.46\\[-1.2pt]{\tiny\color{black!38}$\pm0.14$}} & \cellcolor[HTML]{F7C9CC}\shortstack[c]{85.46\\[-1.2pt]{\tiny\color{black!38}$\pm0.08$}}
& \cellcolor[HTML]{F7C9CC}\shortstack[c]{71.68\\[-1.2pt]{\tiny\color{black!38}$\pm0.13$}} & \cellcolor[HTML]{D3E1F6}\shortstack[c]{74.17\\[-1.2pt]{\tiny\color{black!38}$\pm0.12$}} & \cellcolor[HTML]{F7C9CC}\shortstack[c]{83.68\\[-1.2pt]{\tiny\color{black!38}$\pm0.17$}} & \cellcolor[HTML]{D3E1F6}\shortstack[c]{95.65\\[-1.2pt]{\tiny\color{black!38}$\pm0.09$}} & \cellcolor[HTML]{F7C9CC}\shortstack[c]{81.29\\[-1.2pt]{\tiny\color{black!38}$\pm0.06$}}
& \cellcolor[HTML]{F7C9CC}\shortstack[c]{97.33\\[-1.2pt]{\tiny\color{black!38}$\pm0.08$}} & \cellcolor[HTML]{F7C9CC}\shortstack[c]{84.06\\[-1.2pt]{\tiny\color{black!38}$\pm0.06$}} & \cellcolor[HTML]{F7C9CC}\shortstack[c]{84.27\\[-1.2pt]{\tiny\color{black!38}$\pm0.14$}} & \cellcolor[HTML]{F7C9CC}\shortstack[c]{96.27\\[-1.2pt]{\tiny\color{black!38}$\pm0.06$}} & \cellcolor[HTML]{F7C9CC}\shortstack[c]{90.48\\[-1.2pt]{\tiny\color{black!38}$\pm0.05$}}
\\

\bottomrule
\end{tabular}%
}
\vspace{-4mm}
\end{table*}

Using the shared frozen Gaussian components and stage-specific occupancy
$\boldsymbol{\pi}_s$, we define
\begin{equation}
\label{eq:var_stage_density}
q_s(\mathbf{u})
=
\sum_{k=1}^{K_t}
\pi_{s,k}
\mathcal{N}\!\left(
\mathbf{u};
\boldsymbol{\mu}_k,
\operatorname{diag}(\mathbf{v}_k)
\right).
\end{equation}
The stages therefore share Gaussian components and differ only in authentic
component occupancy.

For any probe sample $x$, its anchor and current negative log-likelihoods are
\begin{equation}
\label{eq:var_nll}
\begin{aligned}
\ell_s^{\mathrm{anc}}(x)
&= -\log q_s\!\left(
\mathbf{u}_{\bar{\theta},\bar{\psi}}(x)
\right),\\
\ell_s^{\mathrm{cur}}(x)
&= -\log q_s\!\left(
\mathbf{u}_{\theta,\psi}(x)
\right).
\end{aligned}
\end{equation}
Let $\nu_s$ and $\kappa_s$ denote the robust center and scale of the anchor
negative log-likelihoods on $\mathcal{P}^{\mathrm{a}}_s$, estimated using
median absolute deviation~\cite{MARKATOU199749}. They normalize the following density-based vulnerability indicators.

{\color{myblue}\triRight}
\textit{Authentic-density degradation $D_s$.}
We measure whether authentic representations move toward lower-density regions
of the stage-specific reference:
\begin{equation}
\label{eq:var_drift}
D_s
=
\frac{1}{|\mathcal{P}^{\mathrm{a}}_s|}
\sum_{x\in\mathcal{P}^{\mathrm{a}}_s}
g\!\left(
\frac{
\left[
\ell_s^{\mathrm{cur}}(x)
-
\ell_s^{\mathrm{anc}}(x)
\right]_{+}
}{
\kappa_s
}
\right),
\end{equation}
where $[\cdot]_{+}=\max(\cdot,0)$ and $g(\cdot)$ smoothly bounds the
normalized likelihood degradation.

{\color{myblue}\triRight}
\textit{Forged-boundary confusion $B_s$.}
A forged sample is vulnerable when it both resembles the authentic reference
and receives an insufficient forged-class margin. Let
$m_{\theta,\phi}(x)=[\mathbf{o}_{\theta,\phi}(x)]_{1}-[\mathbf{o}_{\theta,\phi}(x)]_{0}$ denote this margin. We define
\begin{equation}
\label{eq:var_confusion}
B_s =
\frac{1}{|\mathcal{P}^{\mathrm{f}}_s|}
\sum_{x\in\mathcal{P}^{\mathrm{f}}_s}
\sigma\!\bigl(
\tfrac{\nu_s-\ell_s^{\mathrm{cur}}(x)}
{\kappa_s\tau_{\mathrm{den}}}
\bigr)
\sigma\!\bigl(
-\tfrac{m_{\theta,\phi}(x)}{\tau_{\mathrm{mar}}}
\bigr),
\end{equation}
where $\sigma(\cdot)$ is the sigmoid function, and
$\tau_{\mathrm{den}}$ and $\tau_{\mathrm{mar}}$ are temperature factors.
The two factors measure authentic-region proximity and insufficient
forged-class confidence, respectively.

{\color{myblue}\triRight}
\textit{Authentic-mode occupancy drift $C_s$.}
VAR further measures changes in the aggregate allocation of authentic
representations across Gaussian modes. The current occupancy profile is
\begin{equation}
\label{eq:var_current_profile}
\widehat{\boldsymbol{\pi}}_s
=
\frac{
\mathbf{1}
+
\sum_{x\in\mathcal{M}^{\mathrm{a}}_{t-1,s}}
\mathbf{r}\!\left(
\mathbf{u}_{\theta,\psi}(x)
\right)
}{
K_t
+
|\mathcal{M}^{\mathrm{a}}_{t-1,s}|
}.
\end{equation}
We compare the anchor and current profiles using the Hellinger distance:
\begin{equation}
\label{eq:var_occupancy}
C_s
=
\frac{1}{\sqrt{2}}
\left\|
\sqrt{\boldsymbol{\pi}_s}
-
\sqrt{\widehat{\boldsymbol{\pi}}_s}
\right\|_2.
\end{equation}

Together, $D_s$, $B_s$, and $C_s$ capture authentic-density degradation,
forged-boundary confusion, and authentic-mode redistribution. The stage
vulnerability score is
\begin{equation}
\label{eq:var_raw_risk}
R_s
=
\operatorname{clip}\!\left(
\alpha_{\mathrm{drift}}D_s
+
\alpha_{\mathrm{conf}}B_s
+
\alpha_{\mathrm{occ}}C_s,
0,
1
\right).
\end{equation}

\noindent\textbf{Vulnerability-Aware Replay Allocation.}
To reduce estimation noise, VAR maintains an exponential moving average
$\widetilde{R}_s$ of historical vulnerability estimates.
The relative vulnerability is then converted into a bounded stage weight:
\begin{equation}
\label{eq:var_weight}
\rho_s =
\operatorname{clip}\!\Bigl(
1+\eta\bigl[
\widetilde{R}_s-\tfrac{1}{t}
\sum_{j=0}^{t-1}\widetilde{R}_j
\bigr]_{+},
\rho_{\min},\rho_{\max}
\Bigr).
\end{equation}
Thus, stages with above-average vulnerability receive stronger replay supervision, while the bounded weighting prevents a single stage from dominating optimization. The final VAR objective is a weighted aggregation of historical replay losses: \begin{equation} \label{eq:var_loss} \mathcal{L}_{\mathrm{VAR}} = \frac{ \sum_{s=0}^{t-1} \rho_s \mathcal{L}_{\mathrm{rep}}^{(s)} }{ \sum_{s=0}^{t-1}\rho_s }. \end{equation} 
Here, $\mathcal{L}_{\mathrm{rep}}^{(s)}$ denotes the classification
loss on replay samples from stage $s$.
VAR statistics and weights are detached from gradient computation.
VAR only reweights replay classification; replay sampling,
memory replacement, GMM fitting, and GRASP remain unchanged.

\subsection{Training and Inference}
\label{sec:training_inference}

At stage $t>0$, the complete training objective is
\begin{equation}
\label{eq:arise_objective}
\mathcal{L}_{\mathrm{ARISE}}
=
\mathcal{L}_{\mathrm{cur}}
+
\mathcal{L}_{\mathrm{VAR}}
+
\mathcal{L}_{\mathrm{GRASP}},
\end{equation}
where $\mathcal{L}_{\mathrm{cur}}$ is the binary classification loss on the current-stage samples. At the initial stage, no historical replay or authentic reference is available, so only $\mathcal{L}_{\mathrm{cur}}$ is optimized.

During an incremental update, replay samples pass through both the current and frozen detectors. The current detector produces replay logits for $\mathcal{L}_{\mathrm{VAR}}$ and projected representations for $\mathcal{L}_{\mathrm{GRASP}}$. The frozen detector supplies the corresponding historical anchors and the coordinate system of the authentic GMM. The GMM and stage target profiles stay fixed within a stage, whereas VAR risks and replay weights are updated periodically during optimization. After all epochs of stage $t$, samples from $\mathcal{D}^{\mathrm{tr}}_t$ are inserted into the replay memory by the standard reservoir rule.

\begin{table*}[t]
\centering
\caption{
Cross-benchmark results on Weibo~\cite{Weibo} and
FineFake~\cite{FineFake} (\%).
Results are averaged over three independent runs.
\textit{Avg.} is the mean over the four incremental tasks.
The best and second-best results in each column are shown in
\textbf{bold} and \underline{underlined}, respectively.
}
\vspace{-0.3cm}
\label{tab:cross_benchmark}

\setlength{\tabcolsep}{3.0pt}
\renewcommand{\arraystretch}{1.08}

\resizebox{0.95\textwidth}{!}{%
\begin{tabular}{
    l
    *{10}{c}
    !{\color{DividerGray}\vrule width 0.7pt}
    *{10}{c}
}
\toprule

\multirow{3}{*}{Method}
& \multicolumn{10}{c!{\color{DividerGray}\vrule width 0.7pt}}
  {Weibo~\cite{Weibo}}
& \multicolumn{10}{c}{FineFake~\cite{FineFake}} \\

\cmidrule(lr){2-11}
\cmidrule(lr){12-21}

& \multicolumn{2}{c}{T1}
& \multicolumn{2}{c}{T2}
& \multicolumn{2}{c}{T3}
& \multicolumn{2}{c}{T4}
& \multicolumn{2}{c!{\color{DividerGray}\vrule width 0.7pt}}{Avg.}
& \multicolumn{2}{c}{T1}
& \multicolumn{2}{c}{T2}
& \multicolumn{2}{c}{T3}
& \multicolumn{2}{c}{T4}
& \multicolumn{2}{c}{Avg.} \\

& Acc. & F1
& Acc. & F1
& Acc. & F1
& Acc. & F1
& Acc. & F1
& Acc. & F1
& Acc. & F1
& Acc. & F1
& Acc. & F1
& Acc. & F1 \\

\midrule

\rowcolor{GroupGray}
\multicolumn{21}{l}{\textit{Architecture-based Methods}} \\

MoE-CLIP~\cite{yu2024moeclip}
& 92.44 & 92.26
& 91.16 & 90.40
& 92.53 & 92.01
& 94.65 & 94.45
& 92.70 & 92.28
& 73.58 & 65.00
& 61.00 & 58.93
& 60.46 & 53.07
& 87.69 & 76.48
& 70.68 & 63.37 \\

D-MoLE~\cite{ge2025dynamic}
& 92.68 & 92.09
& 92.49 & 92.11
& 94.17 & 94.05
& 97.65 & 96.83
& 94.25 & 93.77
& \second{76.69} & 73.19
& 71.75 & 70.21
& 66.06 & 60.95
& 86.00 & \best{78.80}
& 75.13 & \second{70.79} \\

DAEDCMD~\cite{wang2025daedcmd}
& 93.01 & 92.12
& \second{94.79} & 92.68
& 93.77 & 93.41
& \best{100.00} & \best{100.00}
& 95.39 & 94.55
& 75.82 & 73.57
& \second{72.38} & 70.26
& 68.23 & \best{63.93}
& 88.61 & 75.22
& \second{76.26} & 70.75 \\

\midrule

\rowcolor{GroupGray}
\multicolumn{21}{l}{\textit{Replay-based Methods}} \\

iCaRL~\cite{rebuffi2017icarl}
& 89.43 & 88.65
& 90.80 & 90.28
& 92.76 & 92.17
& 91.78 & 91.67
& 91.19 & 90.69
& 68.29 & 63.46
& 53.71 & 53.43
& 59.90 & 59.15
& 82.29 & 76.19
& 66.05 & 63.06 \\

LPR~\cite{yoo2024lpr}
& \second{94.16} & \second{94.10}
& 94.18 & \second{94.12}
& \second{95.29} & \second{95.11}
& \best{100.00} & \best{100.00}
& \second{95.91} & \second{95.83}
& 71.45 & 68.92
& 72.14 & \second{72.08}
& 67.18 & 60.74
& \second{89.47} & 73.09
& 75.06 & 68.71 \\

aL-SAR~\cite{seo2025alsar}
& 92.12 & 92.09
& 93.66 & 93.29
& 92.43 & 92.16
& 98.76 & 98.43
& 94.24 & 93.99
& 74.18 & 72.83
& 59.12 & 59.01
& \best{70.19} & \second{61.31}
& 87.04 & 76.12
& 72.63 & 67.32 \\

\midrule

\rowcolor{GroupGray}
\multicolumn{21}{l}{\textit{Regularization-based Methods}} \\

MAS~\cite{aljundi2018mas}
& 88.89 & 88.16
& 89.90 & 88.60
& 90.91 & 90.54
& 94.23 & 94.21
& 90.98 & 90.38
& 72.43 & 64.59
& 64.81 & 62.38
& 60.68 & 57.47
& 84.92 & 74.37
& 70.71 & 64.70 \\

C-Flat~\cite{bian2024cflat}
& 93.90 & 93.78
& 93.75 & 93.08
& 92.41 & 92.16
& \second{99.79} & \second{99.75}
& 94.96 & 94.69
& 75.85 & \best{75.17}
& 71.54 & 69.18
& 65.58 & 58.60
& 87.28 & 74.38
& 75.06 & 69.33 \\

ZAF~\cite{gao2024zaf}
& 92.32 & 92.08
& 93.49 & 92.67
& 94.99 & 93.34
& \best{100.00} & \best{100.00}
& 95.20 & 94.52
& 73.71 & 71.04
& 72.16 & 71.28
& 68.62 & 60.71
& 88.44 & 76.72
& 75.73 & 69.94 \\

\midrule

\rowcolor{OursBlue}
\textbf{\method{} (Ours)}
& \best{94.70} & \best{94.51}
& \best{95.79} & \best{95.63}
& \best{95.77} & \best{95.68}
& \best{100.00} & \best{100.00}
& \best{96.57} & \best{96.46}
& \best{76.82} & \second{74.10}
& \best{73.96} & \best{73.46}
& \second{69.25} & 60.47
& \best{89.78} & \second{77.99}
& \best{77.45} & \best{71.51} \\

\bottomrule
\end{tabular}%
}
\end{table*}

\begin{figure*}[!t]
    \centering

    \begin{minipage}[t]{0.3\textwidth}
        \vspace{0pt}
        \centering
        \includegraphics[
            height=\tripanelheight,
            keepaspectratio
        ]{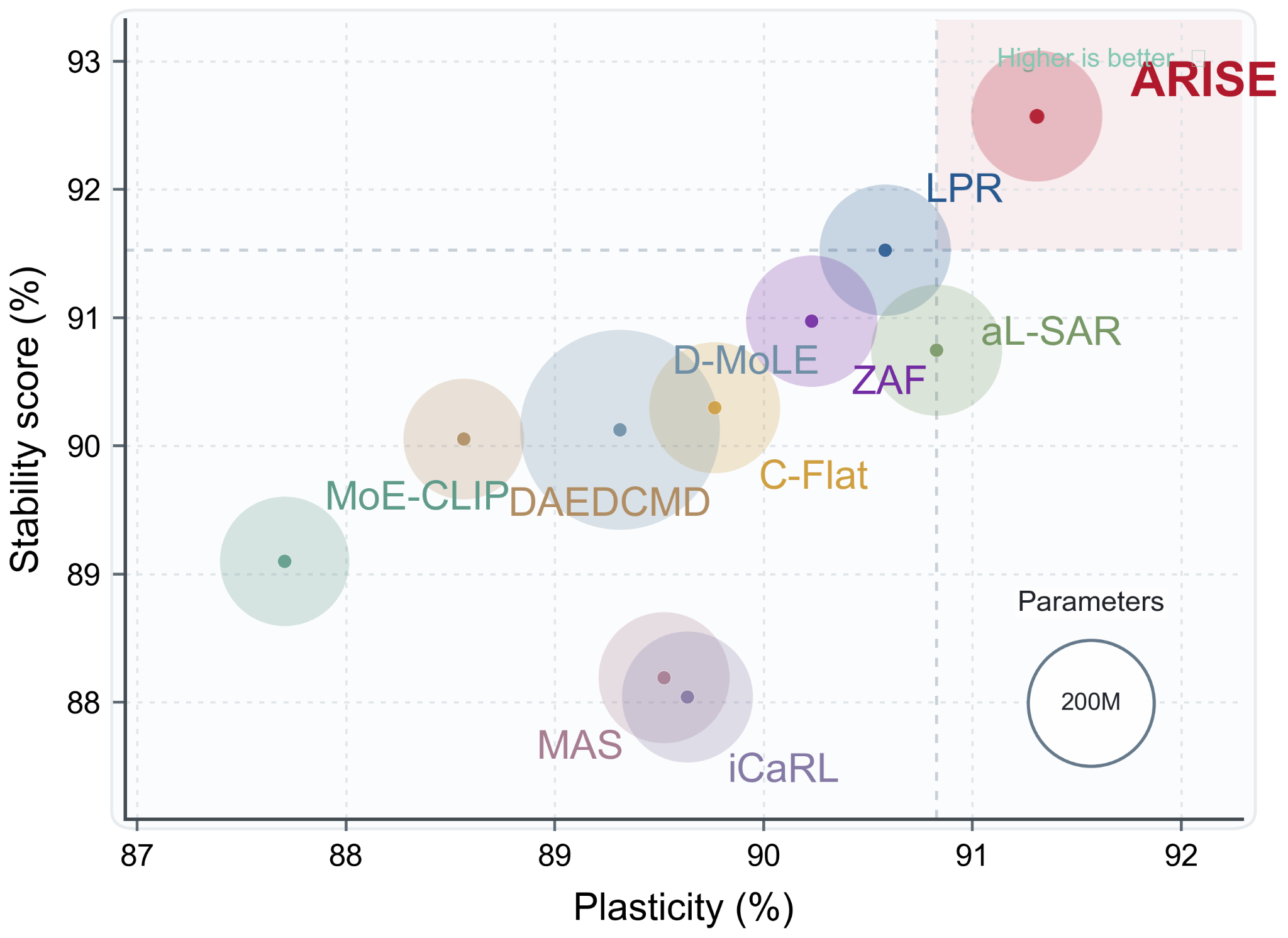}

        {\normalsize
        \textbf{(a)} Plasticity--stability trade-off
        \strut}
    \end{minipage}
    \hfill
    \begin{minipage}[t]{0.30\textwidth}
        \vspace{0pt}
        \centering
        \includegraphics[
            height=\tripanelheight,
            keepaspectratio
        ]{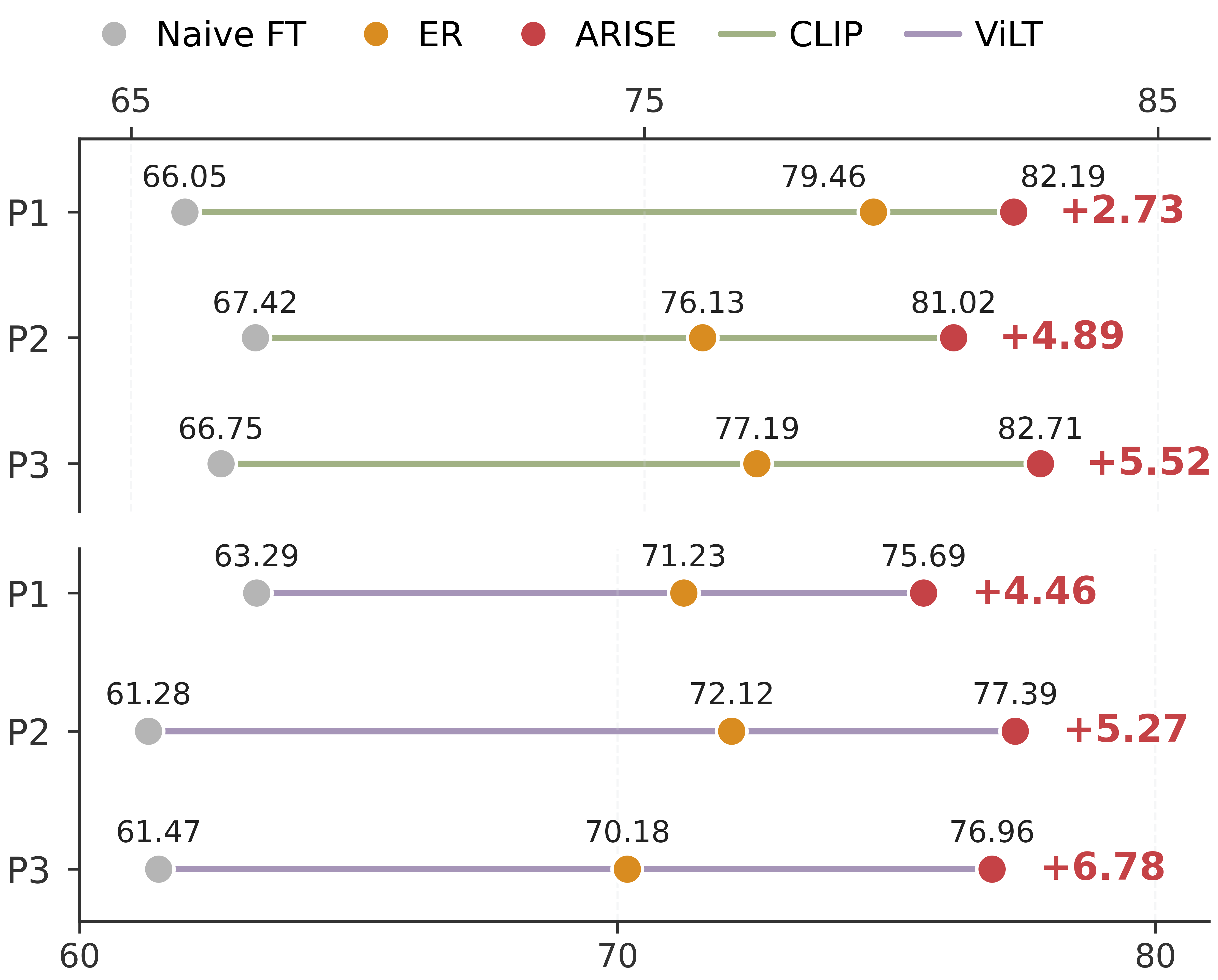}

        {\normalsize
        \textbf{(b)} Plug-and-play Accuracy
        \strut}
    \end{minipage}
    \hfill
    \begin{minipage}[t]{0.3\textwidth}
        \vspace{0pt}
        \centering
        \includegraphics[
            height=\tripanelheight,
            keepaspectratio
        ]{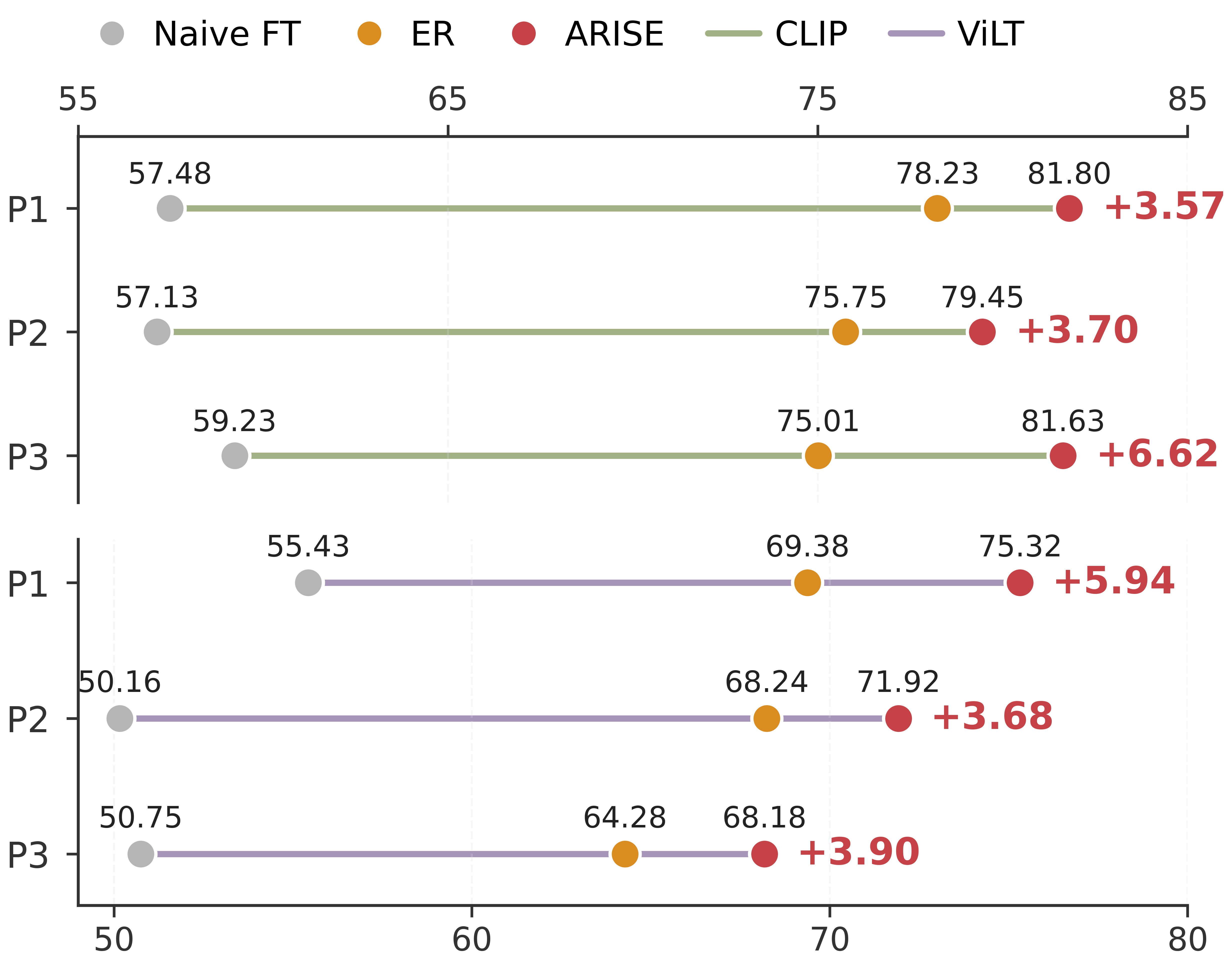}

        {\normalsize
        \textbf{(c)} Plug-and-play Macro-F1
        \strut}
    \end{minipage}

    \vspace{-3mm}
    \caption{
    Overall effectiveness and plug-and-play generalization of \method{}. \textbf{(a)} Plasticity--stability trade-off across continual learning methods; bubble size indicates parameter count. \textbf{(b)--(c)} Accuracy and Macro-F1 when plugged into CLIP and ViLT backbones.
    }
    \label{fig:tradeoff_and_plug_and_play}

    \par\vspace{2mm}

    \begingroup
    \captionof{table}{
    Ablations of \method{}, averaged over three runs. $\Delta$ denotes the mean ACC gain
    over ER across P1--P3 (pp).
    VAR and GRASP are disabled in (b) and (c), respectively.
    Higher is better except for AF.
    Best and second-best results are \textbf{bolded} and \underline{underlined}.
    }
    \label{tab:three_ablation}

    \setlength{\tabcolsep}{1.7pt}
    \renewcommand{\arraystretch}{1.0}

    \def\abheadstrut{\rule[-0.75ex]{0pt}{3.05ex}}
    \def\abrowstrut{\rule[-0.75ex]{0pt}{2.95ex}}

    \def\absecond#1{\smash{\underline{#1}}}

    \def\abfullrowshade{%
        \makebox[0pt][l]{%
            \raisebox{-0.75ex}[0pt][0pt]{%
                \color{OursBlue}\rule{\linewidth}{2.95ex}%
            }%
        }%
    }

    \resizebox{\textwidth}{!}{%
    \begin{minipage}{1.15\textwidth}
    \noindent
    \begin{minipage}[t][3.70cm][t]{0.30\linewidth}
    \vspace{0pt}
    \centering
    \fontsize{7.2}{8.4}\selectfont

    \begin{tabular*}{\linewidth}{
        @{\extracolsep{\fill}}
        l
        cc
        rrr
        r
        @{}
    }
    \multicolumn{1}{c}{
        \abheadstrut
        \multirow{2}{*}{Variant}
    }
    & \multirow{2}{*}{G}
    & \multirow{2}{*}{V}
    & \multicolumn{3}{c}{Avg. ACC (\%) $\uparrow$}
    & \multirow{2}{*}{$\Delta$ (pp)}
    \\

    \abheadstrut
    & & & P1 & P2 & P3 &
    \\

    \midrule[0.9pt]

    \abrowstrut
    Base (ER)
    & --
    & --
    & 81.70
    & 77.67
    & 83.57
    & --
    \\

    \abrowstrut
    Base + GRASP
    & $\checkmark$
    & --
    & \absecond{83.44}
    & \absecond{80.14}
    & \absecond{88.93}
    & \absecond{+3.19}
    \\

    \abrowstrut
    Base + VAR
    & --
    & $\checkmark$
    & 82.12
    & 79.13
    & 88.29
    & +2.20
    \\

    \abrowstrut
    \abfullrowshade
    \textbf{\method{}}
    & $\checkmark$
    & $\checkmark$
    & \textbf{85.37}
    & \textbf{81.22}
    & \textbf{90.47}
    & \textbf{+4.71}
    \\
    \end{tabular*}

    \vspace{8pt}
    \normalsize
    \textbf{(a)} Component Ablation
    \vfill

    \end{minipage}%
    \hfill%
    \begin{minipage}[t][3.70cm][t]{0.315\linewidth}
    \vspace{0pt}
    \centering
    \fontsize{7.2}{8.4}\selectfont

    \begin{tabular*}{\linewidth}{
        @{\extracolsep{\fill}}
        l
        rr
        rr
        @{}
    }
    \multicolumn{1}{c}{
        \abheadstrut
        \multirow{2}{*}{Variant}
    }
    & \multicolumn{2}{c}{Continual Scores}
    & \multicolumn{2}{c}{Protocol Avg. (\%)}
    \\

    \abheadstrut
    & Plasticity $\uparrow$
    & Stability $\uparrow$
    & ACC $\uparrow$
    & F1 $\uparrow$
    \\

    \midrule[0.9pt]

    \abrowstrut
    Iso. Anchor
    & 88.23
    & \absecond{92.39}
    & 82.46
    & 80.56
    \\

    \abrowstrut
    Global Precision
    & \absecond{89.57}
    & 91.56
    & 83.13
    & 81.76
    \\

    \abrowstrut
    Hard Assignment
    & 89.21
    & 92.13
    & \absecond{83.80}
    & \absecond{82.44}
    \\

    \rowcolor{OursBlue}
    \abrowstrut
    \textbf{Full GRASP}
    & \textbf{89.89}
    & \textbf{92.47}
    & \textbf{84.17}
    & \textbf{82.93}
    \\
    \end{tabular*}

    \vspace{8pt}
    \normalsize
    \textbf{(b)} GRASP Ablation
    \vfill

    \end{minipage}%
    \hfill%
    \begin{minipage}[t][3.70cm][t]{0.35\linewidth}
    \vspace{0pt}
    \centering
    \fontsize{7.2}{8.4}\selectfont

    \begin{tabular*}{\linewidth}{
        @{\extracolsep{\fill}}
        l
        rr
        rr
        @{}
    }
    \multicolumn{1}{c}{
        \abheadstrut
        \multirow{2}{*}{Variant}
    }
    & \multicolumn{2}{c}{Continual Scores}
    & \multicolumn{2}{c}{Protocol Avg. (\%)}
    \\

    \abheadstrut
    & AF $\downarrow$
    & Plasticity $\uparrow$
    & ACC $\uparrow$
    & F1 $\uparrow$
    \\

    \midrule[0.9pt]

    \abrowstrut
    ER
    & 16.33
    & \textbf{92.36}
    & 80.98
    & 78.78
    \\

    \abrowstrut
    Geometry ($D_s+C_s$)
    & \absecond{12.09}
    & 91.97
    & \absecond{82.54}
    & \absecond{80.89}
    \\

    \abrowstrut
    Boundary ($B_s$)
    & 14.13
    & \absecond{92.25}
    & 81.89
    & 80.43
    \\

    \rowcolor{OursBlue}
    \abrowstrut
    \textbf{Full VAR ($D_s+B_s+C_s$)}
    & \textbf{11.73}
    & 92.08
    & \textbf{83.18}
    & \textbf{81.61}
    \\
    \end{tabular*}

    \vspace{8pt}
    \normalsize
    \textbf{(c)} VAR Ablation
    \vfill

    \end{minipage}%
    \end{minipage}%
    }
    \endgroup

    \vspace{-10mm}
\end{figure*}


At inference, prediction uses only the multimodal backbone
$f_{\theta}$ and the binary classification head $h_{\mathrm{cls},\phi}$.
The authenticity projection, frozen anchor detector, GRASP,
VAR, and replay memory are
excluded from the inference path.
Therefore, \method{} introduces no stage-specific prediction branch and no
additional inference burden.

%% file: 06_experiments.tex
\section{Experiments}
\label{sec:experiments}

We initialize the backbone of \method{} with publicly available
pretrained Florence-2~\cite{xiao2023florence2} weights and use
a 2,000-sample replay memory.
Replay- and regularization-based baselines share this backbone,
while architecture-based methods retain their native backbones.

\subsection{Main Results on \bench{}}
\label{sec:main_results}

In Table~\ref{tab:main_results}, we compare \method{} with representative
continual-learning methods under the joint-modality (P1),
text-side (P2), and image-side (P3) attack-evolution protocols.
We also include three reference baselines: Joint Training
with access to the complete attack history, Naive Fine-tuning
without replay or continual-learning constraints, and
Experience Replay with stored historical samples.

Zero-shot MLLMs are competitive on the initial tasks of P1
and P2, where direct cross-modal inconsistencies are common.
Their accuracy is substantially lower on later tasks and
varies across image-side attacks in P3.
These results show that the evaluated MLLMs, under zero-shot
prompting, remain less effective than specialized
continual-learning methods across these attack protocols.

Among continual-learning methods, \method{} achieves the
highest average accuracy on all three protocols:
85.46\%, 81.29\%, and 90.48\% on P1--P3, respectively.
It exceeds LPR by 1.63, 1.95, and 1.49 percentage points
and achieves the best accuracy on most individual tasks.
These results support its effectiveness in learning emerging
attacks while retaining performance on previously seen ones.

\subsection{Generalization Evaluation}
\label{sec:generalization}

\noindent\textbf{Cross-benchmark generalization.}
We further evaluate \method{} on Weibo~\cite{Weibo} and
FineFake~\cite{FineFake} using the continual evaluation setup
of~\cite{wang2025daedcmd}. We organize Weibo into chronological
stages by event and split FineFake by publication year.
As shown in Table~\ref{tab:cross_benchmark}, \method{} achieves the
highest average accuracy and F1 on both datasets, with
96.57\%/96.46\% on Weibo and 77.45\%/71.51\% on FineFake.
These results show that \method{} also maintains strong performance under
event-driven and temporal distribution shifts beyond \bench{}.

\noindent\textbf{Backbone transferability.}
We further assess whether \method{} depends on its default Florence-2~\cite{xiao2023florence2} backbone by
integrating it with CLIP~\cite{CLIP} and ViLT~\cite{ViLT}.
As shown in Figs.~\ref{fig:tradeoff_and_plug_and_play}b--c, compared
with standard experience replay, \method{}
consistently improves Accuracy by 2.73--6.78 percentage points and
Macro-F1 by 3.57--6.62 points across all protocols and both backbones.
These improvements demonstrate that \method{} is backbone-agnostic and can be integrated with different
multimodal architectures.

\subsection{Stability--Plasticity Analysis}
\label{sec:tradeoff}

Beyond average accuracy, we examine how \method{} balances
adaptation to emerging attacks and retention of prior knowledge.
We measure \emph{Plasticity} as the average task accuracy
immediately after each task is first learned, and
\emph{Stability} as 100 minus the mean peak-to-final accuracy
drop on earlier tasks.

As shown in Fig.~\ref{fig:tradeoff_and_plug_and_play}(a),
\method{} lies in the upper-right region of the trade-off space,
reaching 91.31\% Plasticity and 92.57\% Stability on P1.
These results show that \method{} adapts to emerging attacks
while retaining performance on previously learned ones.


\subsection{Ablation Studies}
\label{sec:ablation}

In Table~\ref{tab:three_ablation}, we ablate GRASP, VAR, and their key designs.
Average Forgetting (AF) measures the mean drop from each old task's best
to its final performance; lower is better.

\noindent\textbf{Component contributions.}
We compare standard ER with the addition of GRASP, VAR, and their complete
combination in Table~\ref{tab:three_ablation}a.
GRASP improves the averaged Accuracy over P1--P3 by 3.19 percentage
points, while VAR provides a 2.20-point gain.
Combining the two components improves all three protocols and yields an
overall gain of 4.71 points over ER.
These results show that representation-level anchoring and
vulnerability-aware replay provide complementary benefits.

\noindent\textbf{GRASP design.}
We isolate two designs in Table~\ref{tab:three_ablation}b:
component-specific anisotropic precision and responsibility-aware
anchoring.
\textit{Iso. Anchor} replaces the learned precision with the identity
matrix, constraining all representation directions equally.
\textit{Global Precision} keeps anisotropic weighting but shares one
precision estimate across authentic modes.
\textit{Hard Assignment} uses only the most likely Gaussian component
per sample, discarding soft responsibilities.
Full GRASP achieves the highest Plasticity, Stability, and average
Accuracy (89.89\%, 92.47\%, and 84.17\%).

\noindent\textbf{VAR design.}
In Table~\ref{tab:three_ablation}c, we decompose the vulnerability signals used
for replay allocation.
The geometry variant $(D_s+C_s)$ combines authentic-density degradation $D_s$ and
authentic-mode occupancy drift $C_s$, thereby identifying historical stages
whose authentic reference structure is becoming poorly preserved.
The boundary variant $(B_s)$ measures forged-to-authentic boundary confusion
and emphasizes stages whose real--fake decision boundary has become
ambiguous.
Uniform ER uses none of these signals, whereas full VAR combines $D_s$, $B_s$,
and $C_s$.

The geometry signals reduce AF more effectively than the boundary signal
alone, while the boundary signal better preserves Plasticity.
Combining all three signals reduces AF from 16.33\% to 11.73\% and improves
average Accuracy to 83.18\%, with only a 0.28-point decrease
in Plasticity.
VAR therefore strengthens historical stability without materially impairing
the acquisition of emerging attacks.

%% file: 08_conclusion.tex
\section{Conclusion}
\label{sec:conclusion}

We introduced attack-incremental continual multimodal forgery detection, where
a fixed-label detector must learn emerging attacks while retaining previously
acquired forensic knowledge. We developed \bench{} with complementary
joint-modality, text-side, and image-side evolution protocols, and proposed
\method{}, which uses reusable authentic geometry as a compact historical
reference. GRASP provides responsibility-aware anisotropic stabilization,
while VAR emphasizes vulnerable historical stages during replay. Experiments
on \bench{}, two public benchmarks, and multiple backbones show consistent
improvements in final performance, retention, and the stability--plasticity
balance. Since all auxiliary components are used only during continual
updates, \method{} adds no inference-time computation. These results support
continual adaptation as an effective complement to zero-shot generalization
under evolving forgery attacks.

%% file: main.bib
@String(ECCV= {Eur. Conf. Comput. Vis.})

@String(AAAI = {AAAI})

@String(ECCV  = {ECCV})

@inproceedings{wang2018eann,
  author        = {Wang, Yaqing and Ma, Fenglong and Jin, Zhiwei and Yuan, Ye and Xun, Guangxu and Jha, Kishlay and Su, Lu and Gao, Jing},
  title         = {{EANN}: Event Adversarial Neural Networks for Multi-Modal Fake News Detection},
  booktitle     = {Proceedings of the 24th ACM SIGKDD International Conference on Knowledge Discovery \& Data Mining},
  pages         = {849--857},
  year          = {2018},
  publisher     = {Association for Computing Machinery},
  address       = {New York, NY, USA},
  doi           = {10.1145/3219819.3219903}
}

@article{aneja2021cosmos,
  author        = {Aneja, Shivangi and Bregler, Chris and Nie{\ss}ner, Matthias},
  title         = {{COSMOS}: Catching Out-of-Context Image Misuse Using Self-Supervised Learning},
  journal       = {Proceedings of the AAAI Conference on Artificial Intelligence},
  volume        = {37},
  number        = {12},
  pages         = {14084--14092},
  year          = {2023},
  doi           = {10.1609/aaai.v37i12.26648}
}

@inproceedings{ge2025dynamic,
  author        = {Ge, Chendi and Wang, Xin and Zhang, Zeyang and Chen, Hong and Fan, Jiapei and Huang, Longtao and Xue, Hui and Zhu, Wenwu},
  title         = {Dynamic Mixture of Curriculum {LoRA} Experts for Continual Multimodal Instruction Tuning},
  booktitle     = {Proceedings of the 42nd International Conference on Machine Learning},
  series        = {Proceedings of Machine Learning Research},
  volume        = {267},
  pages         = {19011--19033},
  year          = {2025},
  publisher     = {PMLR},
  address       = {Vancouver, Canada},
  url           = {https://proceedings.mlr.press/v267/ge25d.html}
}

@inproceedings{zhang2024asap,
  author        = {Zhang, Zhenxing and Wang, Yaxiong and Cheng, Lechao and Zhong, Zhun and Guo, Dan and Wang, Meng},
  title         = {{ASAP}: Advancing Semantic Alignment Promotes Multi-Modal Manipulation Detecting and Grounding},
  booktitle     = {Proceedings of the IEEE/CVF Conference on Computer Vision and Pattern Recognition},
  pages         = {4005--4014},
  year          = {2025},
  publisher     = {IEEE},
  address       = {Los Alamitos, CA, USA}
}

@inproceedings{shen2025samm,
  author        = {Shen, Jinjie and Wang, Yaxiong and Cheng, Lechao and Pu, Nan and Zhong, Zhun},
  title         = {Beyond Artificial Misalignment: Detecting and Grounding Semantic-Coordinated Multimodal Manipulations},
  booktitle     = {Proceedings of the 33rd ACM International Conference on Multimedia},
  pages         = {11308--11317},
  year          = {2025},
  publisher     = {Association for Computing Machinery},
  address       = {New York, NY, USA},
  doi           = {10.1145/3746027.3754951}
}

@inproceedings{zhang2025mdsm,
  author        = {Zhang, Yuchen and Wang, Yaxiong and Wu, Yujiao and Wu, Lianwei and Zhu, Li and Zheng, Zhedong},
  title         = {The Coherence Trap: When {MLLM}-Crafted Narratives Exploit Manipulated Visual Contexts},
  booktitle     = {Proceedings of the IEEE/CVF Conference on Computer Vision and Pattern Recognition},
  pages         = {8760--8769},
  month         = jun,
  year          = {2026},
  publisher     = {IEEE},
  address       = {Los Alamitos, CA, USA},
  doi           = {10.48550/arXiv.2505.17476},
  url           = {https://arxiv.org/abs/2505.17476}
}

@inproceedings{zhang2026reform,
  author        = {Zhang, Yuchen and Wang, Yaxiong and Han, Kecheng and Wu, Yujiao and Wu, Lianwei and Zhu, Li and Zheng, Zhedong},
  title         = {Cultivating Forensic Reasoning for Generalizable Multimodal Manipulation Detection},
  booktitle     = {Proceedings of the 64th Annual Meeting of the Association for Computational Linguistics (Volume 1: Long Papers)},
  pages         = {28545--28560},
  month         = {July},
  year          = {2026},
  publisher     = {Association for Computational Linguistics},
  address       = {San Diego, California, United States},
  doi           = {10.18653/v1/2026.acl-long.1316}
}

@inproceedings{xiao2023florence2,
  author        = {Xiao, Bin and
              Wu, Haiping and
              Xu, Weijian and
              Dai, Xiyang and
              Hu, Houdong and
              Lu, Yumao and
              Zeng, Michael and
              Liu, Ce and
              Yuan, Lu},
  title         = {{Florence-2}: Advancing a Unified Representation for a Variety of Vision Tasks},
  booktitle     = {Proceedings of the IEEE/CVF Conference on Computer Vision and Pattern Recognition},
  pages         = {4818--4829},
  year          = {2024},
  publisher     = {IEEE},
  address       = {Los Alamitos, CA, USA}
}

@inproceedings{wang2025daedcmd,
  author        = {Wang, Bing and Li, Ximing and Ye, Mengzhe and Li, Changchun and Fu, Bo and Qu, Jianfeng and Wu, Lin Yuanbo},
  title         = {Remember Past, Anticipate Future: Learning Continual Multimodal Misinformation Detectors},
  booktitle     = {Proceedings of the 33rd ACM International Conference on Multimedia},
  pages         = {5490--5498},
  year          = {2025},
  publisher     = {Association for Computing Machinery},
  address       = {New York, NY, USA},
  doi           = {10.1145/3746027.3754759}
}

@misc{Zhou2026REVEAL,
  author        = {Zhou, Jun and Hu, Bingwen and Wang, Yaxiong and Zheng, Zhedong and Wang, Yongzhen and Zhang, Yuchen and Liu, Ping},
  title         = {{REVEAL}: Reference-Grounded Reasoning for Multimodal Manipulation Detection},
  month         = may,
  year          = {2026},
  doi           = {10.48550/arXiv.2605.28459},
  eprint        = {2605.28459},
  archivePrefix = {arXiv},
  primaryClass  = {cs.CV},
  url           = {https://arxiv.org/abs/2605.28459}
}

@misc{openai2026gpt54mini,
  author        = {{OpenAI}},
  title         = {Introducing {GPT-5.4} Mini and Nano},
  month         = mar,
  year          = {2026},
  url           = {https://openai.com/index/introducing-gpt-5-4-mini-and-nano/},
  howpublished  = {OpenAI}
}

@techreport{googledeepmind2025gemini3flash,
  author        = {{Google DeepMind}},
  title         = {{Gemini 3 Flash}: Model Card},
  month         = dec,
  year          = {2025},
  institution   = {Google DeepMind},
  url           = {https://storage.googleapis.com/deepmind-media/Model-Cards/Gemini-3-Flash-Model-Card.pdf}
}

@misc{qwen2026qwen35,
  author        = {{Qwen Team}},
  title         = {{Qwen3.5}: Towards Native Multimodal Agents},
  month         = feb,
  year          = {2026},
  url           = {https://qwen.ai/blog?id=qwen3.5}
}

@misc{wang2025internvl35,
  author        = {Wang, Weiyun and
              Gao, Zhangwei and
              Gu, Lixin and
              Pu, Hengjun and
              Cui, Long and
              Wei, Xingguang and
              Liu, Zhaoyang and
              Jing, Linglin and
              Ye, Shenglong and
              Shao, Jie and
              Wang, Zhaokai and
              Chen, Zhe and
              Zhang, Hongjie and
              Yang, Ganlin and
              Wang, Haomin and
              Wei, Qi and
              Yin, Jinhui and
              Li, Wenhao and
              Cui, Erfei and
              Chen, Guanzhou and
              Ding, Zichen and
              Tian, Changyao and
              Wu, Zhenyu and
              Xie, Jingjing and
              Li, Zehao and
              Yang, Bowen and
              Duan, Yuchen and
              Wang, Xuehui and
              Li, Songze and
              Zhao, Xiangyu and
              Duan, Haodong and
              Deng, Nianchen and
              Fu, Bin and
              He, Yinan and
              Wang, Yi and
              He, Conghui and
              Shi, Botian and
              He, Junjun and
              Xiong, Yingtong and
              Lv, Han and
              Wu, Lijun and
              Shao, Wenqi and
              Zhang, Kaipeng and
              Deng, Huipeng and
              Qi, Biqing and
              Ge, Jiaye and
              Guo, Qipeng and
              Zhang, Wenwei and
              Ouyang, Wanli and
              Wang, Limin and
              Dou, Min and
              Zhu, Xizhou and
              Lu, Tong and
              Lin, Dahua and
              Dai, Jifeng and
              Zhou, Bowen and
              Su, Weijie and
              Chen, Kai and
              Qiao, Yu and
              Wang, Wenhai and
              Luo, Gen},
  title         = {{InternVL3.5}: Advancing Open-Source Multimodal Models in Versatility, Reasoning, and Efficiency},
  year          = {2025},
  doi           = {10.48550/arXiv.2508.18265},
  eprint        = {2508.18265},
  archivePrefix = {arXiv},
  primaryClass  = {cs.CV},
  url           = {https://arxiv.org/abs/2508.18265}
}

@misc{liu2024llavanext,
  author        = {Liu, Haotian and Li, Chunyuan and Li, Yuheng and Li, Bo and Zhang, Yuanhan and Shen, Sheng and Lee, Yong Jae},
  title         = {{LLaVA-NeXT}: Improved Reasoning, OCR, and World Knowledge},
  month         = jan,
  year          = {2024},
  url           = {https://llava-vl.github.io/blog/2024-01-30-llava-next/}
}

@article{FineFake,
  author        = {Ziyi Zhou and Xiaoming Zhang and Litian Zhang and Jiacheng Liu and Erik Cambria and Chaozhuo Li},
  title         = {FineFake: A knowledge-enriched dataset for fine-grained multi-domain fake news detection},
  journal       = {Information Fusion},
  volume        = {132},
  pages         = {104253},
  month         = aug,
  year          = {2026},
  publisher     = {Elsevier},
  issn          = {1566-2535},
  doi           = {10.1016/j.inffus.2026.104253},
  url           = {https://www.sciencedirect.com/science/article/pii/S1566253526001326}
}

@inproceedings{Weibo,
  author        = {Jin, Zhiwei and Cao, Juan and Guo, Han and Zhang, Yongdong and Luo, Jiebo},
  title         = {Multimodal Fusion with Recurrent Neural Networks for Rumor Detection on Microblogs},
  booktitle     = {Proceedings of the 25th ACM International Conference on Multimedia},
  series        = {MM '17},
  pages         = {795--816},
  numpages      = {22},
  year          = {2017},
  publisher     = {Association for Computing Machinery},
  address       = {New York, NY, USA},
  isbn          = {9781450349062},
  doi           = {10.1145/3123266.3123454},
  url           = {https://doi.org/10.1145/3123266.3123454},
  location      = {Mountain View, California, USA}
}

@incollection{MARKATOU199749,
  author        = {Markatou, M. and Ronchetti, E.},
  title         = {Robust Inference: The Approach Based on Influence Functions},
  booktitle     = {Robust Inference},
  series        = {Handbook of Statistics},
  volume        = {15},
  pages         = {49--75},
  year          = {1997},
  publisher     = {Elsevier},
  address       = {Amsterdam, The Netherlands},
  editor        = {Maddala, G. S. and Rao, C. R.},
  doi           = {10.1016/S0169-7161(97)15005-2}
}

@inproceedings{CVPR_DGM4,
  author        = {Shao, Rui and Wu, Tianxing and Liu, Ziwei},
  title         = {Detecting and Grounding Multi-Modal Media Manipulation},
  booktitle     = {Proceedings of the IEEE/CVF Conference on Computer Vision and Pattern Recognition},
  pages         = {6904--6913},
  year          = {2023},
  publisher     = {IEEE},
  address       = {Los Alamitos, CA, USA},
  doi           = {10.1109/CVPR52729.2023.00667}
}

@article{Olan2024_fakeNewsImpact,
  author        = {Olan, Femi and Jayawickrama, Uchitha and Arakpogun, Emmanuel Ogiemwonyi and Suklan, Jana and Liu, Shaofeng},
  title         = {Fake news on Social Media: the Impact on Society},
  journal       = {Information Systems Frontiers},
  volume        = {26},
  number        = {2},
  pages         = {443--458},
  year          = {2024},
  issn          = {1572-9419},
  doi           = {10.1007/s10796-022-10242-z}
}

@article{delange2022survey,
  author        = {De Lange, Matthias and
              Aljundi, Rahaf and
              Masana, Marc and
              Parisot, Sarah and
              Jia, Xu and
              Leonardis, Ale{\v{s}} and
              Slabaugh, Gregory and
              Tuytelaars, Tinne},
  title         = {A Continual Learning Survey: Defying Forgetting in Classification Tasks},
  journal       = {IEEE Transactions on Pattern Analysis and Machine Intelligence},
  volume        = {44},
  number        = {7},
  pages         = {3366--3385},
  year          = {2022},
  doi           = {10.1109/TPAMI.2021.3057446}
}

@inproceedings{ER_replay,
  author        = {Rolnick, David and Ahuja, Arun and Schwarz, Jonathan and Lillicrap, Timothy P. and Wayne, Gregory},
  title         = {Experience Replay for Continual Learning},
  booktitle     = {Advances in Neural Information Processing Systems},
  volume        = {32},
  pages         = {348--358},
  year          = {2019},
  publisher     = {Curran Associates, Inc.},
  address       = {Red Hook, NY, USA}
}

@inproceedings{shu2019defend,
  author        = {Shu, Kai and Cui, Limeng and Wang, Suhang and Lee, Dongwon and Liu, Huan},
  title         = {{dEFEND}: Explainable Fake News Detection},
  booktitle     = {Proceedings of the 25th ACM SIGKDD International Conference on Knowledge Discovery \& Data Mining},
  pages         = {395--405},
  year          = {2019},
  publisher     = {Association for Computing Machinery},
  address       = {New York, NY, USA},
  doi           = {10.1145/3292500.3330935}
}

@inproceedings{ye2021theoretical,
  author        = {Ye, Haotian and Xie, Chuanlong and Cai, Tianle and Li, Ruichen and Li, Zhenguo and Wang, Liwei},
  title         = {Towards a Theoretical Framework of Out-of-Distribution Generalization},
  booktitle     = {Advances in Neural Information Processing Systems},
  volume        = {34},
  pages         = {23519--23531},
  year          = {2021},
  publisher     = {Curran Associates, Inc.},
  address       = {Red Hook, NY, USA}
}

@inproceedings{CLIP,
  author        = {Radford, Alec and
              Kim, Jong Wook and
              Hallacy, Chris and
              Ramesh, Aditya and
              Goh, Gabriel and
              Agarwal, Sandhini and
              Sastry, Girish and
              Askell, Amanda and
              Mishkin, Pamela and
              Clark, Jack and
              Krueger, Gretchen and
              Sutskever, Ilya},
  title         = {Learning Transferable Visual Models from Natural Language Supervision},
  booktitle     = {Proceedings of the 38th International Conference on Machine Learning},
  series        = {Proceedings of Machine Learning Research},
  volume        = {139},
  pages         = {8748--8763},
  year          = {2021},
  publisher     = {PMLR},
  address       = {Virtual Event},
  url           = {https://proceedings.mlr.press/v139/radford21a.html}
}

@inproceedings{ViLT,
  author        = {Kim, Wonjae and Son, Bokyung and Kim, Ildoo},
  title         = {{ViLT}: Vision-and-Language Transformer Without Convolution or Region Supervision},
  booktitle     = {Proceedings of the 38th International Conference on Machine Learning},
  series        = {Proceedings of Machine Learning Research},
  volume        = {139},
  pages         = {5583--5594},
  year          = {2021},
  publisher     = {PMLR},
  address       = {Virtual Event},
  url           = {https://proceedings.mlr.press/v139/kim21k.html}
}

@inproceedings{yu2024moeclip,
  author        = {Yu, Jiazuo and Zhuge, Yunzhi and Zhang, Lu and Hu, Ping and Wang, Dong and Lu, Huchuan and He, You},
  title         = {Boosting Continual Learning of Vision-Language Models via Mixture-of-Experts Adapters},
  booktitle     = {Proceedings of the IEEE/CVF Conference on Computer Vision and Pattern Recognition},
  pages         = {23219--23230},
  year          = {2024},
  publisher     = {IEEE},
  address       = {Los Alamitos, CA, USA},
  doi           = {10.1109/CVPR52733.2024.02191}
}

@inproceedings{rebuffi2017icarl,
  author        = {Rebuffi, Sylvestre-Alvise and Kolesnikov, Alexander and Sperl, Georg and Lampert, Christoph H.},
  title         = {{iCaRL}: Incremental Classifier and Representation Learning},
  booktitle     = {Proceedings of the IEEE Conference on Computer Vision and Pattern Recognition},
  pages         = {5533--5542},
  year          = {2017},
  publisher     = {IEEE},
  address       = {Los Alamitos, CA, USA},
  doi           = {10.1109/CVPR.2017.587}
}

@inproceedings{yoo2024lpr,
  author        = {Yoo, Jinsoo and Liu, Yunpeng and Wood, Frank and Pleiss, Geoff},
  title         = {Layerwise Proximal Replay: A Proximal Point Method for Online Continual Learning},
  booktitle     = {Proceedings of the 41st International Conference on Machine Learning},
  series        = {Proceedings of Machine Learning Research},
  volume        = {235},
  pages         = {57199--57216},
  year          = {2024},
  publisher     = {PMLR},
  address       = {Vienna, Austria},
  url           = {https://proceedings.mlr.press/v235/yoo24a.html}
}

@inproceedings{seo2025alsar,
  author        = {Seo, Minhyuk and Koh, Hyunseo and Choi, Jonghyun},
  title         = {Budgeted Online Continual Learning by Adaptive Layer Freezing and Frequency-based Sampling},
  booktitle     = {The Thirteenth International Conference on Learning Representations},
  year          = {2025},
  publisher     = {OpenReview.net},
  address       = {Singapore},
  url           = {https://openreview.net/forum?id=dOAkHmsjRX}
}

@inproceedings{aljundi2018mas,
  author        = {Aljundi, Rahaf and Babiloni, Francesca and Elhoseiny, Mohamed and Rohrbach, Marcus and Tuytelaars, Tinne},
  title         = {Memory Aware Synapses: Learning What (Not) to Forget},
  booktitle     = {Computer Vision -- ECCV 2018},
  series        = {Lecture Notes in Computer Science},
  volume        = {11207},
  pages         = {144--161},
  year          = {2018},
  publisher     = {Springer},
  address       = {Cham, Switzerland},
  doi           = {10.1007/978-3-030-01219-9_9}
}

@inproceedings{bian2024cflat,
  author        = {Bian, Ang and
              Li, Wei and
              Yuan, Hangjie and
              Yu, Chengrong and
              Wang, Mang and
              Zhao, Zixiang and
              Lu, Aojun and
              Ji, Pengliang and
              Feng, Tao},
  title         = {Make Continual Learning Stronger via {C-Flat}},
  booktitle     = {Advances in Neural Information Processing Systems},
  volume        = {37},
  pages         = {7608--7630},
  year          = {2024},
  publisher     = {Curran Associates, Inc.},
  address       = {Red Hook, NY, USA},
  doi           = {10.52202/079017-0244}
}

@inproceedings{gao2024zaf,
  author        = {Gao, Zijian and Zhang, Xingxing and Xu, Kele and Mao, Xinjun and Wang, Huaimin},
  title         = {Stabilizing Zero-Shot Prediction: A Novel Antidote to Forgetting in Continual Vision-Language Tasks},
  booktitle     = {Advances in Neural Information Processing Systems},
  volume        = {37},
  pages         = {128462--128488},
  year          = {2024},
  publisher     = {Curran Associates, Inc.},
  address       = {Red Hook, NY, USA},
  doi           = {10.52202/079017-4080}
}
